\documentclass[journal]{IEEEtran}
\usepackage{times}
\usepackage{helvet}
\usepackage{courier}

\usepackage{microtype}
\usepackage[american]{babel}
\usepackage{makecell}
\usepackage{algorithm}
\usepackage{algorithmic}
\usepackage{amsmath,amssymb}
\usepackage{bm}
\usepackage{amsfonts}
\usepackage{amsthm}
\usepackage{multirow}
\usepackage{graphicx}
\usepackage{graphics}
\usepackage{color}
\usepackage{url}
\usepackage{array}
\usepackage{verbatim}

\usepackage{caption}

\newcommand{\x}{\textbf{x}}

\newcommand{\f}{\textbf{f}}
\newcommand{\h}{\textbf{h}}

\begin{document}
%
\title{Learning with Interpretable Structure \\from Gated RNN}
%
%
%

\author{Bo-Jian Hou,
        and~Zhi-Hua Zhou,~\IEEEmembership{Fellow,~IEEE}
\thanks{ All authors are with the National Key Laboratory for Novel Software
Technology, Nanjing University, Nanjing 210023, China. E-mail: \{houbj,zhouzh\}@lamda.nju.edu.cn.}}

\markboth{SUBMITTED TO THE SPECIAL ISSUE ON STRUCTURED MULTI-OUTPUT LEARNING: MODELLING, ALGORITHM, THEORY AND APPLICATIONS}%
{Shell \MakeLowercase{\textit{et al.}}: Bare Demo of IEEEtran.cls for IEEE Journals}


\maketitle

\begin{abstract}


The interpretability of deep learning models has raised extended attention these years. It will be beneficial if we can learn an interpretable structure from deep learning models. In this paper, we focus on Recurrent Neural Networks~(RNNs) especially gated RNNs whose inner mechanism is still not clearly understood. We find that Finite State Automaton~(FSA) that processes sequential data has more interpretable inner mechanism according to the definition of interpretability and can be learned from RNNs as the interpretable structure. We propose two methods to learn FSA from RNN based on two different clustering methods. With the learned FSA and via experiments on artificial and real datasets, we find that FSA is more trustable than the RNN from which it learned, which gives FSA a chance to substitute RNNs in applications involving humans' lives or dangerous facilities. Besides, we analyze how the number of gates affects the performance of RNN. Our result suggests that gate in RNN is important but the less the better, which could be a guidance to design other RNNs. Finally, we observe that the FSA learned from RNN gives semantic aggregated states and its transition graph shows us a very interesting vision of how RNNs intrinsically handle text classification tasks.

\end{abstract}


%
\IEEEpeerreviewmaketitle

\section{Introduction}
\label{section:Introduction}
In the last several years, great advances have been produced in Recurrent Neural Networks (RNNs), and especially those with gates (gated RNNs, such as MGU with one gate~\cite{DBLP:journals/corr/ZhouWZZ16}, GRU with two gates~\cite{DBLP:conf/emnlp/ChoMGBBSB14} and LSTM with three gates~\cite{DBLP:journals/neco/HochreiterS97}) have been successfully applied to various tasks, such as speech recognition~\cite{DBLP:journals/spm/X12a}, image caption~\cite{DBLP:conf/cvpr/VinyalsTBE15}, sentiment analysis~\cite{DBLP:conf/emnlp/TangQL15}, etc. However, nowadays, as the inner mechanisms of learning algorithms become more and more complex, their interpretability is also becoming more and more important~\cite{DBLP:journals/corr/Lipton16a,DBLP:journals/corr/KarpathyJL15,DBLP:journals/corr/YosinskiCNFL15,DBLP:conf/aaai/WuHPZ0D18}. For example, Freitas (2013)~\cite{DBLP:journals/sigkdd/Freitas13} stressed that the comprehensibility (interpretability) of a model to the user is important. Concretely, in applications involving human being's lives or dangerous facilities such as nuclear power plant, reliable models are required which should not only give high accuracy but also have explanation why they made the decision. In \emph{learnware}~\cite{DBLP:journals/fcsc/Zhou16a}, Zhou also stressed this point where people usually want to know what have been learned by models, particularly in real tasks where decision reliability is crucial and rigorous judgment by human beings are critical.

It is worthy to note that there are no unified definitions on the interpretability in machine learning. We are satisfied with the definitions provided by Kim et al. (2016)~\cite{DBLP:conf/nips/KimKK16} that \emph{interpretability is the degree to which a human can consistently predict the model's result}. Another definition we appreciate is that \emph{interpretability is the degree to which a human can understand the cause of a decision}~\cite{DBLP:journals/ai/Miller19}. Lipton (2016)~\cite{DBLP:journals/corr/Lipton16a} also discussed a specific form of interpretability known as \emph{human-simulability} where a human-simulatable model is one in which a human user can take in input data with the parameters of the model and in reasonable time steps through every calculation required to produce a prediction or briefly \emph{the one can be simulated by human beings}. These three perspectives of interpretability are actually consistent with each other. That is, if a human can simulate what the model did or can consistently predict the model's result, the human can understand the cause of the model's decision and trust it.

The inner mechanism of gated RNNs is complex due to three factors. One is their recurrent structure inherited from classical RNN~\cite{DBLP:journals/cogsci/Elman90}. Despite that the recurrent structure has shown to be the key in handling sequential data, using the same unit recurrently for different inputs will make human beings confused about the inner process of classification. Another complexity of gated RNNs comes from the gates used on the unit. Although gates can bring the benefits of remembering long memory, the function of gates has not been fully understood; especially how many gates are inherently required for a gated RNN model. Gates also increase the amount of parameters intensively. Thirdly, the inner process of gated RNN is in the form of non-linear numerical calculation, while non-linear operations and numerical vectors could not be directly associated to a concrete meaning for people to understand. Gated RNNs are more like black boxes compared to many other machine learning models, such as SVMs~\cite{DBLP:conf/ecml/Joachims98} or decision trees~\cite{DBLP:journals/ml/Quinlan86}, and there is still no clear understanding of the inner working mechanisms of recurrent neural networks.  In a word, besides successful applications, we also require understanding of the inner mechanisms of gated RNNs.

We realize that besides RNNs, there is another tool capable of processing sequential data, i.e. Finite State Automaton (FSA)~\cite{nla.cat-vn710549}. FSA is composed of finite states and transitions between states. Its state will transit from one to another in response to external sequential inputs. The transition process of FSA is similar to that of RNN when both models accept items from some sequence one by one, and transit between states accordingly. The inner mechanism of FSA is easier to be simulated by human beings since a human can take in input data together with the transition function or the illustration of FSA and in reasonable time steps through every transition required to produce a prediction. This characteristic exactly conforms to the definition of interpretability aforementioned. Thus, FSA is an interpretable model. It would be interesting if we can learn this interpretable structure from gated RNNs. In this way, we may have a chance to use the interpretability of FSA to probe into gated RNNs and find some interesting or even insightful phenomena to help us interpret or understand gated RNNs to some degree. Besides, with this interpreting or understanding, we may have a rule or guidance to design other RNNs. The idea of learning a simple and interpretable structure is inspired by an early work NeC4.5~\cite{DBLP:journals/tkde/ZhouJ04} which learns an understandable decision tree by observing the behavior of the ensemble of several neural networks instead of using the true labels directly. This operation is also adopted later in knowledge distillation~\cite{DBLP:journals/corr/HintonVD15} which is an effective approach and improves the understandability of deep neural networks.


As for how to learn the interpretable structure, i.e., FSA, we are inspired by early studies about rule extraction from neural networks~\cite{DBLP:journals/jcst/Zhou04}, in particular, pioneering works about rule extraction from RNNs~\cite{DBLP:journals/nn/OmlinG96,DBLP:journals/neco/ZengGS93}. They found that hidden states of classical non-gated RNNs tend to form clusters, with which an FSA can be extracted to represent the rules or grammars learned by the corresponding non-gated RNNs. We want to follow this idea to learn FSAs from gated RNNs and to use this interpretable structure to probe into RNNs. We cannot directly extend \cite{DBLP:journals/nn/OmlinG96,DBLP:journals/neco/ZengGS93} to gated RNNs since there are several differences between these two works and ours. First, we do not know whether the tendency to form clusters will also hold for gated RNNs. Besides, they mainly focus on rule extraction or how well the grammar is extracted by FSA while we focus on interpretability. Furthermore, the clustering method they used are based on quantization which may cause the number of clusters grows exponentially as the number of hidden units increases. Thus, it is not possible to directly extend ~\cite{DBLP:journals/nn/OmlinG96,DBLP:journals/neco/ZengGS93} to gated RNNs. Some recent works~\cite{DBLP:journals/neco/WangZOXLG18,wang2018comparative} use k-means~\cite{hartigan1979algorithm} to cluster the hidden states of RNNs, but they still focus on the quality of grammar that FSA learned. Although we use more advanced clustering methods, note that clustering method is not the key problem here since various effective clustering methods have been proposed without the exponential computing problem~\cite{DBLP:books/daglib/p/Berkhin06}. The principles to choose clustering method are the simplicity and efficiency. We believe other clustering methods can also provide similar results.

Learning from multiple data resources~\cite{DBLP:journals/tip/GongTMLKY16,DBLP:conf/aaai/Gong17}, or training several basic models and then combining them~\cite{zhou2012ensemble} usually produce better results. Thus, besides learning only one FSA from RNNs, we also generate multiple FSAs to do ensemble~\cite{zhou2012ensemble}, where the diversity of individuals is crucial for the improvement of performance~\cite{Sun2018,Zhang2019}. Furthermore, single structure may contain limited semantic information, whereas multiple structures might make the semantic information more plentiful and better to understand.


In this paper, we probe into RNNs through learning FSA from them. We first verify that besides RNN without gates, gated RNNs' hidden states also have the natural tendency to form clusters. Then we propose two methods. One is based on the stable and high-efficient clustering methods \emph{k-means++}~\cite{DBLP:conf/soda/ArthurV07}. The other makes use of the observations that hidden states close in the same sequence also tend to be near in geometrical spaces, named as \emph{k-means-x}. We then learn FSA by designing its five necessary elements, i.e., alphabet, a set of states, start state, a set of accepting state and state transitions. We apply our methods on artificial data and real-world data. We find that FSA is more trustable than the RNN from which it learned, which gives FSA a chance to substitute RNNs in applications involving humans' lives or dangerous facilities, etc. Besides, we analyze how the number of gates affects the performance of RNN. Our result suggests that gate in RNN is important but the less the better, which could be a guidance to design other RNNs. Additionally, we observe that the FSA learned from RNN gives semantic aggregated states and its transition graph shows us a very interesting vision of how RNNs intrinsically handle text classification tasks. We also explore the multiple structures and find that multiple FSAs can improve performance and make the semantic information more plentiful.
Overall, our contributions are mainly fourfold:
\begin{enumerate}
\item We propose two effective methods to learn the interpretable structure, i.e., FSA from gated RNNs. The state transitions in RNNs can be well visualized in a simplified FSA graph. Through this graph, people can easily simulate the running process of FSA so as to validate the interpretability of FSA. We also show that FSA is consistent with its RNN from which it learned. In this way, FSA has a chance to substitute its RNN in applications that need trust.
\item By studying the learned FSA from four main versions of RNNs with different number of gates from zero to three (i.e., SRN, MGU, GRU and LSTM), we find that RNNs with a single gate (i.e., MGU) enjoys a more compact hidden state representation (could be clustered into fewer clusters for FSA state definition) than RNNs with no gate or more than one gates. This phenomenon suggests us that gate in RNN is necessary, but RNN with simple gate structure such as MGU is more desirable and this rule can be our guidance to design other RNN models.
\item On real-world sentiment analysis data, the FSA learned from RNN gives semantic aggregated states and its transition graph shows us a very interesting vision of how RNNs intrinsically handle text classification tasks. Concretely speaking, we find that the words leading the same transitions tend to form word class and possess the same sentiment. The word class leading transitions to accepting state contains mainly positive words while the word class leading transitions to rejecting state contains mainly negative words.
\item We learn many FSAs instead of one and leverage ensemble technique to combine these FSAs to enhance all the results we present, which shows multiple structures combined by ensemble have an edge over single structure.
\end{enumerate}


In the following, we are going to introduce background. Then we state our detailed algorithms, followed by experiments with sufficient discussions. Finally, we conclude our work.

\section{Background}
\label{section:Background}
In this section, we introduce one non-gated RNN and three gated RNNs, which will be studied in our paper, followed by introduction on related works.

First, we introduce the classical non-gated RNN. It was proposed in early 90s~\cite{DBLP:journals/cogsci/Elman90} with simple structure which does not possess any gate and is only applied to small scale data. Therefore, we call it Simple RNN (SRN). In general, SRN takes each element of a sequence as an input and combines it with the hidden state from the last time to calculate the current hidden state iteratively. Concretely, at time $t$, we input the $t$-th element of a sequence, saying $\x_t$ into the hidden unit. Then the hidden unit will give the output $\h_t$ based on the current input $\x_t$ and the previous hidden state $\h_{t-1}$ in the following way: 
$\h_t=f(\h_{t-1},\x_t).$ $f$ is usually defined as a linear transformation plus a nonlinear activation, e.g., $\h_t=\tanh(W[\h_{t-1},\x_t]+\textbf{b})$ where the matrix $W$ consists of parameters related to $\h_{t-1}$ and $\x_t$ and $\textbf{b}$ is a bias term. The task of SRN is to learn $W$ and $\textbf{b}$.

However, the data we are facing are growing bigger and bigger, thus we need deeper model~\cite{DBLP:conf/nips/KrizhevskySH12,DBLP:books/daglib/0040158} to tackle this problem. Yet in this situation, SRN will suffer from the vanishing or exploding gradient issue, which makes learning SRN using gradient descent very difficult~\cite{DBLP:journals/neco/HochreiterS97,DBLP:journals/tnn/BengioSF94}. Fortunately, gated RNNs are proposed to solve the gradient issue by introducing various gates to hidden unit to control how information flows in RNN. The two prevailing gated RNNs are Long Short Term Memory~(LSTM)~\cite{DBLP:journals/neco/GersSC00} and Gated Recurrent Unit~(GRU)~\cite{DBLP:conf/emnlp/ChoMGBBSB14}. LSTM has three gates including an input gate controlling adding of new information, a forget gate determining remembering of old information and an output gate deciding outputting of current information. GRU has two gates, an update gate and a reset gate which controls forgetting of old information and adding of new information, respectively, similar to the forget and input gate in LSTM.



The previous models add several gates to one hidden unit, producing many additional parameters to tune and compute, thus may not be efficient enough. To tackle this problem, Zhou et al. (2016)~\cite{DBLP:journals/corr/ZhouWZZ16} produced Minimal Gated Unit~(MGU), which only has a forget gate and has comparable performance with LSTM and GRU. Thus, MGU's structure is simpler, parameters are fewer and training and tuning are faster than the previous mentioned gated RNNs. 

\begin{table}[!t]
\setlength{\abovedisplayskip}{0pt}
\setlength{\belowdisplayskip}{-0.15cm}
	\caption{\small	Summary of three gated RNNs (MGU, GRU and LSTM). The bold letters are vectors. $\sigma$ is the logistic sigmoid function, and $\odot$ is the component-wise product between two vectors.}
	\label{table:modern RNN}
	\centering
	\normalsize
		\begin{tabular}{l}
\rule{8cm}{0.8pt}\\
MGU (minimal gated unit)\\
\vbox{
\begin{eqnarray*}
\text{(gate)}\quad\f_t&=&\sigma(W_f[{\h}_{t-1},\x_t]+\textbf{b}_f)\\
\tilde{\h}_t&=&\mbox{tanh}(W_h[\f_t\odot \h_{t-1},\x_t]+\textbf{b}_h)\\
{\h}_t&=&(1-\f_t)\odot \h_{t-1}+\f_t\cdot \tilde{\h}_t.
\end{eqnarray*}}\\
\rule{8cm}{0.8pt}\\
GRU (gated recurrent unit)\\
\vbox{
\begin{eqnarray*}
\text{(gate)}\quad\textbf{z}_t&=&\sigma(W_z[{\h}_{t-1},\x_t]+\textbf{b}_z)\\
\text{(gate)}\quad\textbf{r}_t&=&\sigma(W_r[{\h}_{t-1},\x_t]+\textbf{b}_r)\\
\tilde{\h}_t&=&\mbox{tanh}(W_h[{\textbf{r}_t\cdot \h_{t-1}},\x_t]+\textbf{b}_h)\\
{\h}_t&=&(1-\textbf{z}_t)\odot \h_{t-1} + \textbf{z}_t\cdot\tilde{\h}_t.
\end{eqnarray*}}\\
\rule{8cm}{0.8pt}\\
LSTM (long short-term memory)\\
\vbox{
\begin{eqnarray*}
\text{(gate)}\quad\f_t&=&\sigma(W_f[{\h}_{t-1},\x_t]+\textbf{b}_f)\\
\text{(gate)}\quad\textbf{i}_t&=&\sigma(W_i[{\h}_{t-1},\x_t]+\textbf{b}_i)\\
\text{(gate)}\ \ \textbf{o}_t&=&\sigma(W_o[{\h}_{t-1},\x_t]+\textbf{b}_o)\\
\tilde{\textbf{c}}_t&=&\mbox{tanh}(W_c[{\h}_{t-1},\x_t]+\textbf{b}_c)\quad\quad\\
\textbf{c}_t&=&\f_t\odot \textbf{c}_{t-1}+\textbf{i}_t\odot\tilde{\textbf{c}}_t\\
{\h}_t&=&\textbf{o}_t\odot \mbox{tanh}(\textbf{c}_t).
\end{eqnarray*}}\\
\rule{8cm}{0.8pt}

		\end{tabular}
\end{table}

The mathematical formalizations of the three gated RNN models including MGU, GRU and LSTM mentioned above are summarized in Table~\ref{table:modern RNN}, in which
\[
\sigma(x)=\frac{1}{1+\exp(-x)}
\]
is the logistic sigmoid function (applied to every component of the vector input) and $\odot$ is the component-wise product between two vectors. All gates in Table~\ref{table:modern RNN} are marked with text ``(gate)", from which we can easily see that MGU has one gate, GRU has two gates and LSTM has three gates.


Recently, two representative efforts on understanding RNNs have been made. Karpathy et al. (2015)~\cite{DBLP:journals/corr/KarpathyJL15} focus on visualizing the input features, using heat map to show activations in text input which reveals that hidden state captures the structure of input text, such as length of lines and quotes. Another work~\cite{DBLP:conf/emnlp/ChoMGBBSB14} cares about the relationship between representations, using t-SNE~\cite{DBLP:journals/jmlr/Maaten09} to visualize the phrase representations learned by an RNN encoder-decoder model. Yet they do not consider the relationship between hidden neurons, thus the inner mechanisms cannot be fully explained. Some other understanding works concern another group of DNN model, i.e., Convolutional Neural Networks (CNNs)~\cite{DBLP:conf/nips/KrizhevskySH12} in aspects of visualizing~\cite{DBLP:journals/corr/YosinskiCNFL15,DBLP:journals/corr/ZhouKLOT14,DBLP:conf/eccv/ZeilerF14,DBLP:conf/isvc/Harley15} and relationship between representations or neurons~\cite{DBLP:journals/tvcg/RauberFFT17,DBLP:journals/tvcg/LiuSLLZL17}. However, due to the sequential property of RNNs, these works on CNNs cannot be used for RNN models. 

Besides, some structured models that provide interpretability is focusing on ``model interpretability" rather than ``human interpretability" that we mentioned in Introduction. For example, SISTA-RNN~\cite{DBLP:journals/corr/WisdomPPA16} is based on an explicit probabilistic model such that its weights are interpretable. OFMHS~\cite{Guo2020} combines factorization machine and hierarchical sparsity together so as to explore the hierarchical structure behind the input variables to compensate the loss of interpretability. However, a human still cannot simulate the running process of them, which does not conform to the definition of interpretability we mentioned in Introduction. Thus we need to give some new insights on interpreting the inner mechanisms of gated RNNs, making the model understandable for human beings. In this paper, we will learn the interpretable structure, i.e., FSA to probe into the gated RNNs and attempt to make some contributions to interpretability. 


\begin{figure*}[!t]
\setlength{\abovecaptionskip}{1mm}
\setlength{\belowcaptionskip}{-0.cm}
\centering 
\small
\begin{minipage}{0.24\linewidth}\centering
  \includegraphics[width=1\textwidth]{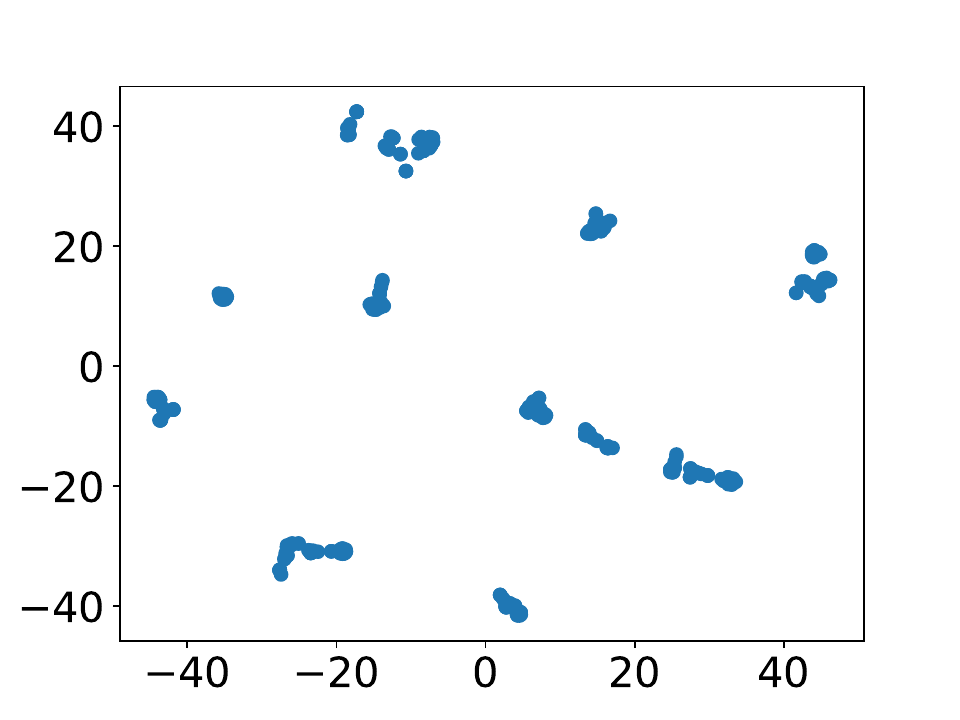}
  \mbox{\footnotesize(a) MGU}
\end{minipage}
\begin{minipage}{0.24\linewidth}\centering
  \includegraphics[width=1\textwidth]{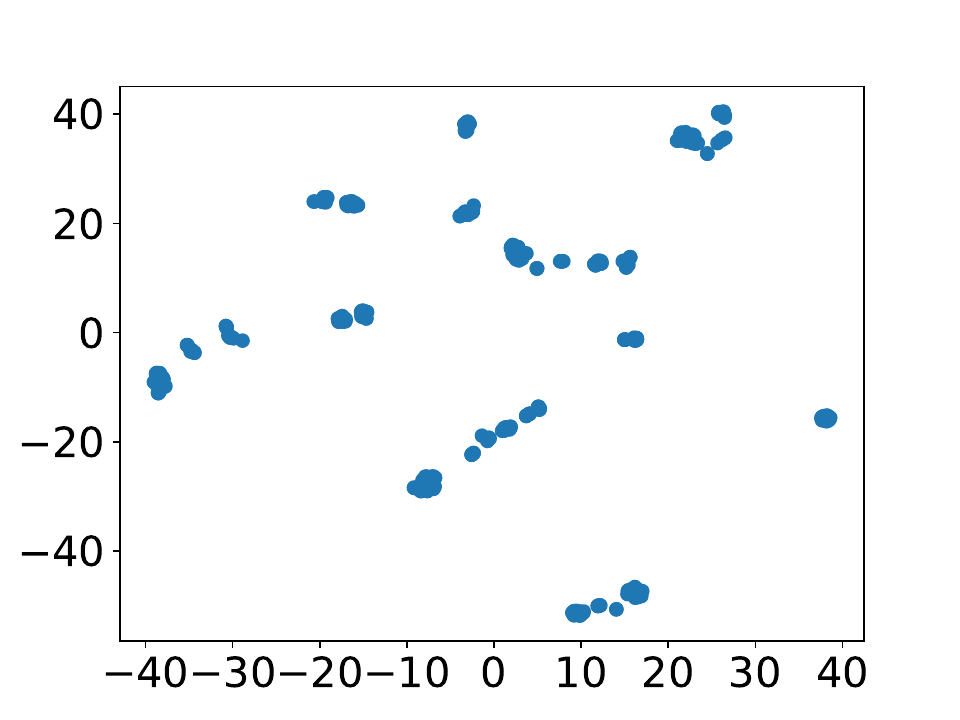}
  \mbox{\footnotesize(b) SRN}
\end{minipage}
\begin{minipage}{0.24\linewidth}\centering
  \includegraphics[width=1\textwidth]{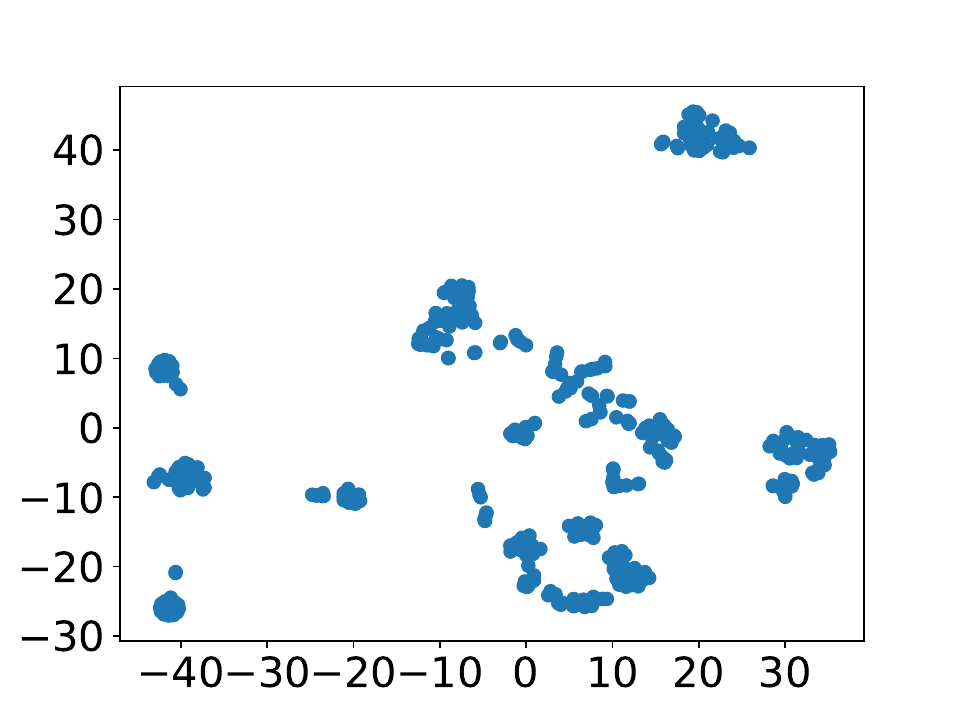}
  \mbox{\footnotesize(c) GRU}
\end{minipage}
\begin{minipage}{0.24\linewidth}\centering
  \includegraphics[width=1\textwidth]{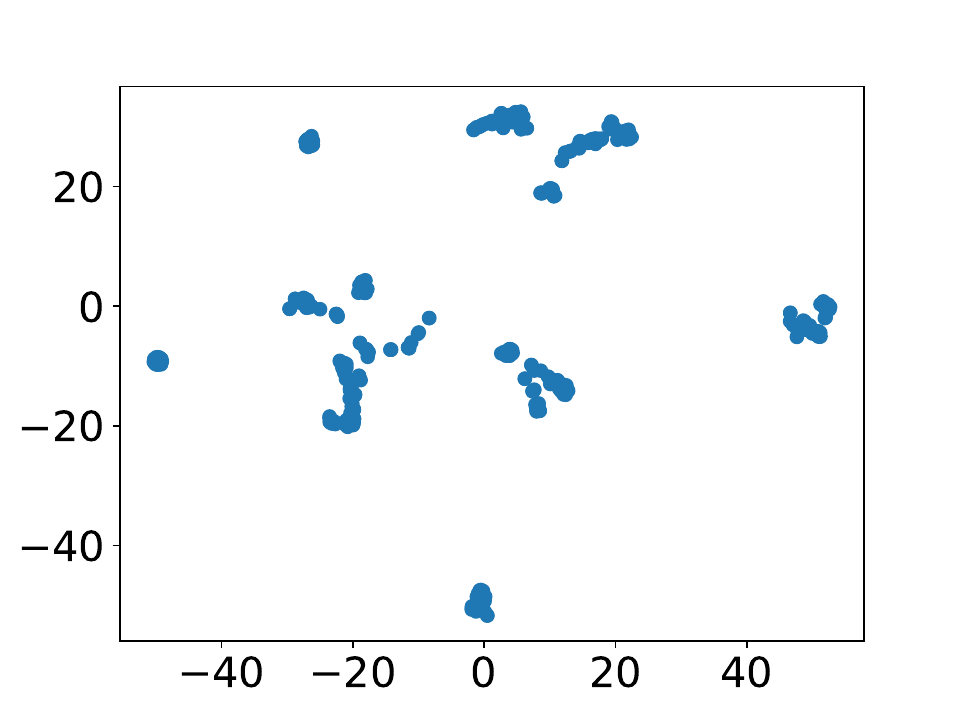}
  \mbox{\footnotesize(d) LSTM}
\end{minipage}
\vspace{2mm}
\caption{\small The hidden state points reduced to two dimensions by t-SNE are plotted. We can see that the hidden state points tend to form cluster for MGU, SRN, GRU and LSTM.}
\label{cluster}
\end{figure*}

\begin{figure*}[!t]
\setlength{\abovecaptionskip}{1mm}
\setlength{\belowcaptionskip}{-0.cm}
\centering 
\small
\includegraphics[width=0.9\textwidth]{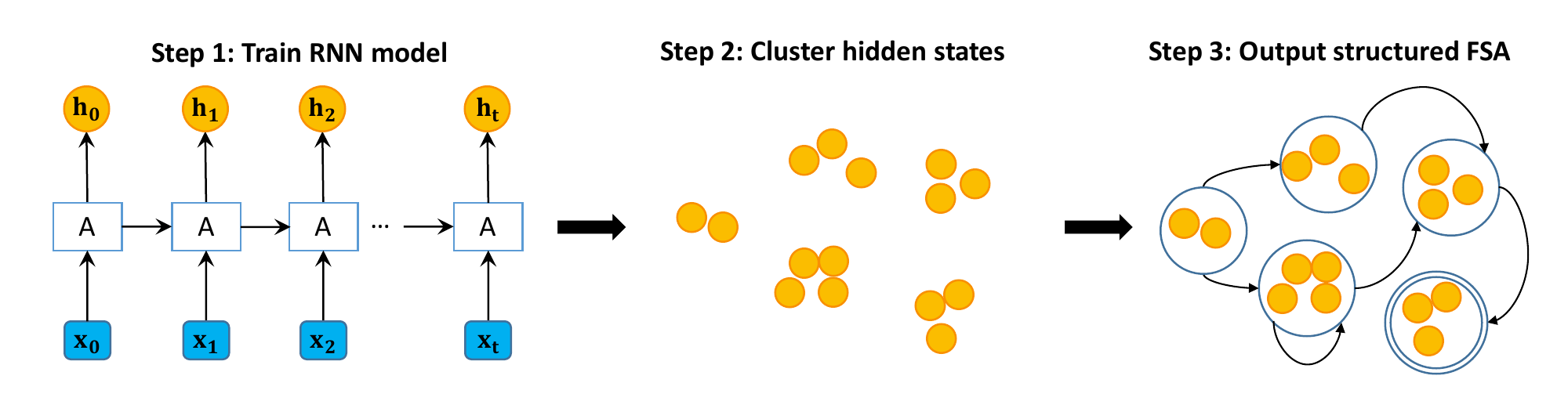}
\caption{\small The illustration of the framework of the proposed algorithm. The yellow circles represent the hidden states denoted as $\h_t$ where $t$ is time step. ``$\mathrm{A}$" is the recurrent unit which receives input $\x_t$ and $\h_{t-1}$ and outputs $\h_t$. The double circle in the structured FSA is the accepting state. Overall, the framework consists of three mains steps, namely, training an RNN model, clustering the RNN's hidden states and outputting the final structured FSA.}
\label{framework}
\end{figure*}

\section{Our Approach}
\label{section: proposed methods}
In this section, we first introduce the intuition and framework, followed by the details of the proposed method including clustering hidden states and learning FSA.

\subsection{Intuition and Framework}

From an intuitive point of view, we consider the following case. First we train an RNN model $\mathcal{R}$ on training data. Then two test sequences $a$ and $b$ are input to $\mathcal{R}$ separately. It is reasonable to observe that if the two subsequences input to $\mathcal{R}$ before time $t_1$ of $a$ and time $t_2$ of $b$ are analogous, the hidden states at time step $t_1$ of $a$ and $t_2$ of $b$ will also resemble each other. We regard a hidden state as a vector or a point. Thus when several sequences are input to RNNs, large amount of hidden state points will accumulate, and similar hidden states is able to form clusters. To validate that, we show the distribution of hidden state points when testing from MGU, SRU, GRU and LSTM respectively in Figure~\ref{cluster}. We set the original dimension of hidden states by 10. Then we use t-Distributed Stochastic Neighbor Embedding~(t-SNE)~\cite{DBLP:journals/jmlr/Maaten09} to reduce the dimension of all 400 hidden state points from 10 to 2 so that we can plot them on the plane.
We already know the fact that the hidden states of SRN tend to form clusters~\cite{DBLP:journals/nn/OmlinG96,DBLP:journals/neco/ZengGS93}. As can be seen from Figure~\ref{cluster}, the clustering results of MGU, GRU and LSTM are similar with that of SRN. Thus, we can infer that the hidden states of MGU, GRU and LSTM also have the property of clustering. Besides, we have 400 points for each RNN models, but we can clearly observe that in Figure~\ref{cluster} (a), 1 (c) and 1 (d), there are only about 11, 9, 9 clusters, respectively, which shows their clustering property.
We assume different clusters will represent different \emph{states}. And \emph{transitions} between states arise when one item of input sequence is read in. Hence the network behaves like a \emph{state automaton}. We assume the states are finite, then we can learn a \emph{Finite State Automaton}~(FSA) from a trained RNN model.



So the overall framework is shown in Figure~\ref{framework}. We firstly train RNNs on training data and then do clustering on all hidden states $H$ corresponding to validation data $V$ and finally learn an FSA with respect to $V$. In the first step of training RNNs, we exploit the same strategy as~\cite{DBLP:journals/corr/ZhouWZZ16} and omit the details here. In the following, we elaborate hidden state clustering and FSA learning steps.

\subsection{Hidden States Clustering}
The first clustering method we consider exploiting is \emph{k-means}~\cite{hartigan1979algorithm}. K-means is to minimize the average squared Euclidean distance of points from their cluster centers, which is efficient, effective and widely used. To obtain a robust result, we use a variant of k-means named as \emph{k-means++}~\cite{DBLP:conf/soda/ArthurV07} by using more robust seeding method to select cluster centers.

Nevertheless, directly using Euclidean distance may not be the best choice. It is reasonable to assume that the hidden state points in the same sequence are more similar, and the hidden state points that are close in time are also near in space. Thus, to consider this characteristic, we concatenate the original hidden state points with extra features that reflect the time closeness. We present an illustration as follows:
\begin{equation*}
\begin{split}
\text{Feature vector of $j$th element in the $i$th sequence}: \\
\overbrace{h_j^1,h_j^2,\ldots,h_j^d,}^{\text{hidden state}}\ \overbrace{\underbrace{0,\ldots,0}_{i-1},j,\underbrace{0,\ldots,0}_{n-i}}^{\text{extra position feature}}, \;\;\;\;\;\;\;\;\;\;
\end{split}
\end{equation*}
where $h_j^d$ means the $d$-th dimension of hidden state point $h_j$. The dimension $n$ of the extra feature is the number of sequences in $V$. Note that each element of a sequence corresponds to a hidden state. For the $j$-th element in the $i$-th sequence, the content in the $i$-th position of the extra feature is $j$. We call the extra feature ``extra position feature". After altering the space, we still use k-means++ to do clustering on the new space. We call this cluster method \emph{k-means-x}.

\vspace{0.5em}
\subsection{Learning FSA}
FSA $M$ is a 5-tuple $M=\left\langle \Sigma, Q, S, F, \delta\right\rangle $ where $\Sigma$ is \emph{alphabet}, meaning the set of the elements appearing in the input sequences, $Q$ is a set of states, $S\in Q$ is the start state, $F\subseteq Q$ is a set of accepting states and $\delta:Q\times\Sigma\rightarrow Q$ defines state transitions in $M$. In order to learn an FSA, we will specify the details of how to design such five elements below.

In our case, we want to learn FSAs (Finite State Automata) from gated RNNs. The alphabet $\Sigma$ is easy to obtain from data. For example, if the data $D$ are sentences consisting of words, then $\Sigma$ is equal to all words in all sentences. So we have 
\begin{equation}
\label{alphabet}
\Sigma=\text{Vocabulary}(D),
\end{equation} 
where $\text{Vocabulary}(D)$ means the vocabulary of $D$.

Every time we input an element from some sequence into RNN, we can get the current hidden state $\h_t$ given the previous hidden state $\h_{t-1}$. This process is similar to that we input a symbol $s$ from alphabet $\Sigma$, and according to the current state and state transitions function $\delta:Q\times\Sigma\rightarrow Q$, we would know which state should be transited to. Thus, we can regard a cluster consisting of several similar hidden state points as a state in FSA. Then, the set of states $Q$ are
\begin{equation}
\label{set of states}
Q = \{\mathcal{C}\ |\ \h\in\mathcal{C}\}\cup \{S\},
\end{equation}
where $\mathcal{C}$ is the cluster of a number of hidden states points $\h$.

We define the start state $S$ by an empty state without any hidden state point because when we input a word into RNN, no previous hidden states are given. Thus the start state $S$ is just a starting symbol. The accepting states $F$ can be determined by the cluster center. Note that each state in FSA is a cluster of hidden state points in RNN. We use the RNN's classifier to classify the cluster center of each state. If the classification result is positive, then the corresponding state is an accepting state, namely,
\begin{equation}
\label{accepting state}
F=\{\mathcal{C}\ |\ \mathcal{R}(\text{cluster center of } \mathcal{C})=1\}
\end{equation}

The fifth element $\delta$ is the most difficult one to obtain among the five elements. We use transition matrix $T\in[|Q|]^{|Q|\times |\Sigma|}$ to represent the state transitions $\delta:Q\times\Sigma\rightarrow Q$ where $|Q|$ means the number of elements in $Q$, $[|Q|]$ means the set of integers ranging from $1$ to $|Q|$ and $|\Sigma|$ means the number of symbols in $\Sigma$. In $T$, each row represents one state (the first row represents the start state $S$, its serial number in $Q$ is $|Q|$), each column represents a symbol $s$ in alphabet. $T(i,j)$ means state $i$ will transit to state $T(i,j)$ when inputting a symbol $s_j$ whose corresponding hidden state point belongs to the $j$-th state. To get a transition matrix $T$, we first need to calculate a matrix $N_s$ for each symbol $s$ in alphabet (e.g. 0 or 1 in binary alphabet), where the $(i,k)$-th entry represents the frequency of jumping from state $i$ to state $k$ given $s$ in all sequences, using the following steps:
\begin{enumerate}
\item indexing every cluster or state, associating each hidden state point to a state in FSA;
\item iterating through all hidden state points, and increasing $N_s(i,k)$ by one when $s$ incurs a transition from state $i$ to state $k$.
\end{enumerate}
As a consequence, $N_s(i,k)$ represents the transition times from state $i$ to $k$ when inputting $s$. In this case, when inputting $s$, state $i$ may transit to several states. We intend to obtain a deterministic FSA for clear illustrating, so we only keep the biggest value which means abandoning the less frequent transitions in each row of $N_s$. Then the transition matrix $T$ can be quickly calculated as follows:
\begin{equation}
\label{transition matrix}
T(i,j)=\mathop{\arg\max}_{k} N_{s_j}(i,k)
\end{equation}
We can draw an illustration of FSA according to $T$ and use $T$ to do classification. When doing classification, the state will keep jumping from one state to another in response to sequentially input symbols, until the end of the sequence. If the final state is an accepting state, the sequence is predicted to be positive by FSA. 

The whole process of learning FSA from RNN is presented in Algorithm~\ref{alg:FSA}. We call our method LISOR~(Learning with Interpretable Structure frOm gated Rnn) and present two concrete algorithms according to different clustering methods. The one based on k-means++ is named as ``LISOR-k" while the other one based on k-means-x is called ``LISOR-x". By utilizing the tendency to form clustering of hidden state points, both LISOR-k and LISOR-x can learn a well generalized FSA from RNN models.

\begin{algorithm}[!t]
	\centering
	\caption{LISOR}
	\label{alg:FSA}
	\begin{algorithmic}[1]
		\REQUIRE
		The number of clusters $k$;\\
		\ENSURE
		An FSA.\\
		\STATE Train an RNN model $\mathcal{R}$ and test on validation data $V$;
		\STATE Record the hidden state point at every time step of every sequence in $V$;
		\STATE Do clustering on the recorded hidden state points $H$;
		\STATE Obtain alphabet $\Sigma$ according to (\ref{alphabet});
		\STATE Obtain set of states according to (\ref{set of states});
		\STATE Obtain accepting states according to (\ref{accepting state});
		\STATE Calculate a matrix $N_s$ for each symbol $s$ in alphabet (e.g. 0 or 1 in binary alphabet); 
		\STATE Generate transition matrix $T$ according to (\ref{transition matrix}).
	\end{algorithmic}
\end{algorithm}

\section{Experiments and Discussions}

\begin{table*}[!t]
\renewcommand\arraystretch{1.2}
\setlength{\abovecaptionskip}{0.cm}
\setlength{\belowcaptionskip}{-0.cm}
	\caption{\small The number of clusters ($n_c$) when the accuracy of FSA learned from four RNNs first achieves 1.0 on task ``0110" by LISOR-k and LISOR-x. Note that the smaller the value is the better for higher efficiency and better interpretability. RNN models trained from different trials are with different initializations on parameters. We can see that on average FSA learned from MGU always possesses the smallest number of clusters when achieving the accuracy 1.0. The number of clusters is 65 means that the FSA's accuracy cannot meet 1.0 when $n_c$ is up to 64 since we set $n_c$ varying from 2 to 64. The smallest number of clusters in each trial and on average are bold.}
	\label{table:accuracy of 0110}
	\vspace{0.8mm}
	\centering
	\small
	\setlength\tabcolsep{10pt}
		\begin{tabular}{c|rrrr|c|rrrr}
\hline
\multicolumn{5}{c|}{LISOR-k} & \multicolumn{5}{c}{LISOR-x}\\
\hline
RNN Type & \makecell[l]{MGU} & \makecell[l]{SRN} & \makecell[l]{GRU} & \makecell[l]{LSTM} & RNN Type & \makecell[l]{MGU} & \makecell[l]{SRN} & \makecell[l]{GRU} & \makecell[l]{LSTM} \\ 
\hline
Trial 1 & \textbf{5} & 13 & 7 & 13 & Trial 1 & \textbf{5} & 13 & 8 & 15 \\

Trial 2 & \textbf{8} & 9 & 25 & 9 & Trial 2 & \textbf{8} & 9 & 65 & 10 \\

Trial 3 & \textbf{6} & 6 & 8 & 12 & Trial 3 & \textbf{6} & \textbf{6} & 8 & 12 \\

Trial 4 & \textbf{5} & \textbf{5} & 8 & 17 & Trial 4 & \textbf{5} & \textbf{5} & 8 & 65 \\

Trial 5 & \textbf{6} & 22 & 9 & 22 & Trial 5 & \textbf{6} & 20 & 9 & 24 \\

Average & \textbf{6} & 11 & 11.2 & 14.6 & Average & \textbf{6} & 10.6 & 19.6 & 25.2\\
\hline
		\end{tabular}
\end{table*}

\begin{table*}[!t]
\renewcommand\arraystretch{1.2}
\setlength{\abovecaptionskip}{0.cm}
\setlength{\belowcaptionskip}{-0.cm}
	\caption{\small The number of clusters ($n_c$) when the accuracy of FSA learned from four RNNs first achieves 0.7 on task ``000" by LISOR-k and LISOR-x. Note that the smaller the value is the better for higher efficiency and better interpretability. RNN models trained from different trials are with different initializations on parameters. We can see that on average FSA learned from MGU always possesses the smallest number of clusters when achieving accuracy 0.7. The number of clusters is 201 means that the FSA's accuracy cannot meet 0.7 when $n_c$ is up to 200 since we set $n_c$ varying from 2 to 200. The smallest number of clusters in each trial and on average are bold.}
	\label{table:accuracy of 000}
	\vspace{0.8mm}
	\centering
	\small
	\setlength\tabcolsep{10pt}
		\begin{tabular}{c|rrrr|c|rrrr}
\hline
\multicolumn{5}{c|}{LISOR-k} & \multicolumn{5}{c}{LISOR-x}\\
\hline
RNN Type & \makecell[l]{MGU} & \makecell[l]{SRN} & \makecell[l]{GRU} & \makecell[l]{LSTM} & RNN Type & \makecell[l]{MGU} & \makecell[l]{SRN} & \makecell[l]{GRU} & \makecell[l]{LSTM} \\ 
\hline
Trial 1 & 38 & 84 & 201 & \textbf{26} & Trial 1 & 31 & 52 & 156 & \textbf{25} \\

Trial 2 & \textbf{6} & 28 & 109 & 72 & Trial 2 & \textbf{6} & 27 & 137 & 60 \\

Trial 3 & \textbf{9} & 28 & 201 & 20 & Trial 3 & \textbf{9} & 18 & 201 & 26 \\

Trial 4 & \textbf{8} & 41 & 85 & 19 & Trial 4 & \textbf{8} & 39 & 91 & 22 \\

Trial 5 & \textbf{7} & 180 & 201 & 22 & Trial 5 & \textbf{7} & 145 & 201 & 39 \\

Average & \textbf{13.6} & 72.2 & 159.4 & 31.8 & Average & \textbf{12.2} & 56.2 & 157.2 & 34.4\\

\hline
		\end{tabular}
\end{table*}

In this section, we conduct experiments on both artificial and real tasks. With experimental results and sufficient discussions, we show how we interpret RNN models via FSAs mainly from the perspective of \emph{human simulating}, \emph{gate effect}, \emph{semantic analysis} and \emph{the advantage of multiple structures}.


\subsection{Artificial Tasks}
In this section, we explore two artificial tasks. 

\subsubsection{Settings}
The first artificial task is to identify sequence 0110 in a group of length-4 sequences which only contain 0 and 1 (task ``0110"). If it is 0110, it will be positive. This is a simple task containing 16 distinct cases. We include 1000 instances in the training sets, with duplicated instances to improve accuracy. We use validation set containing all possible length 4 zero-one sequences without duplication to learn FSAs and randomly generate 100 instances to do testing. 

The second task is to determine whether a sequence contains three consecutive zeros (task ``000"). If it contains consecutive zeros, it will be negative. There is no limitation on the length of sequences, thus the task has infinite instance space and is more difficult than task ``0110". We randomly generate 3000 zero-one training instances whose lengths are also randomly decided. For simplicity, we set the length to be from 5 to 20. We also generate 500 validation and 500 testing instances. 

For both tasks we mainly study the representative RNNs including MGU, SRN, GRU and LSTM mentioned in Section~\ref{section:Background}. For all these four RNN models, we set the dimension of hidden state and the number of hidden layers to be 10 and 3 respectively. We conduct each experiment 5 trials and report the average results.

\subsubsection{Discussions on the Number of Clusters}
\label{subsubsection: number of clusters}
According to Algorithm~\ref{alg:FSA}, in order to learn and visualize an FSA, we need to set the cluster number $k$ or equally, the number of states in FSA. Note that more clusters means each cluster contains less hidden state points. A trivial example is that the number of clusters is equal to the number of hidden state points, and then the state transition in FSA resembles the way that hidden state points transit in RNNs. So the performance of FSA should be close to that of RNNs when $k$ is large enough. Nevertheless, we hope the number of states in FSA to be as small as possible to prevent over-fitting, increase efficiency and reduce complexity so as to be easily simulated by human beings. Thus, achieving high accuracy with small number of clusters is a good characteristic and we are attempting to make the number of clusters as small as possible with guaranteed classification performance. 


\begin{figure*}[!t]
\vspace{-0.8cm}
\setlength{\abovecaptionskip}{1mm}
\setlength{\belowcaptionskip}{-0.cm}
\centering 
\small
\begin{minipage}{0.42\linewidth}\centering
  \vspace{0.4cm}  
  \includegraphics[width=1\textwidth]{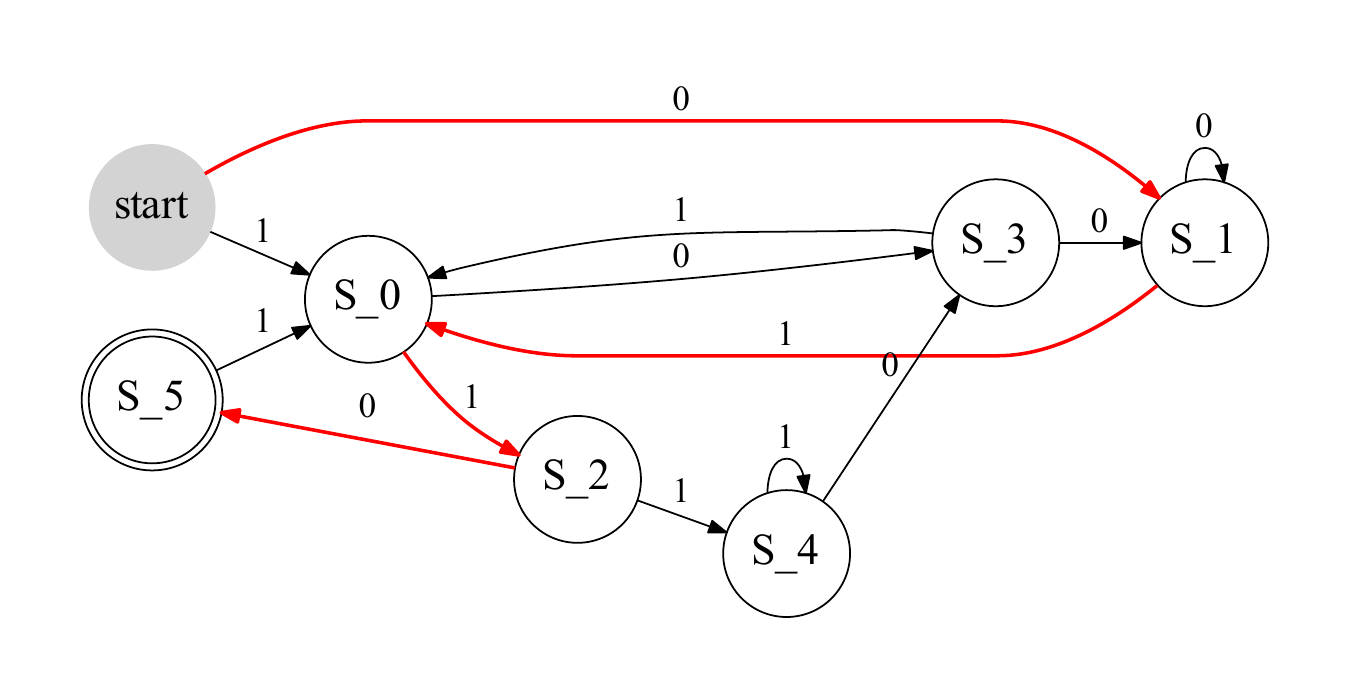}\\
  \mbox{\footnotesize(a) $k=6$ @MGU}
\end{minipage}
\begin{minipage}{0.42\linewidth}\centering
  \includegraphics[width=1\textwidth]{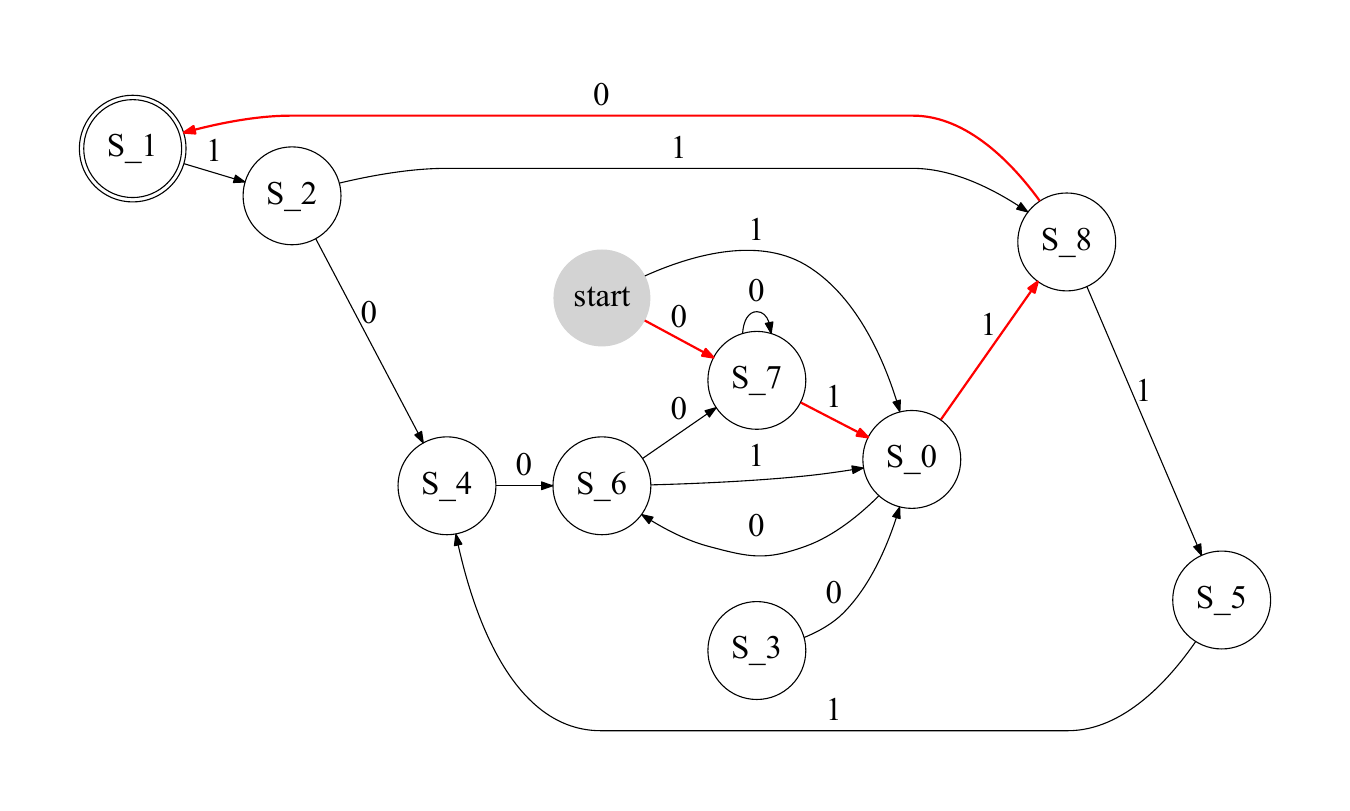}\\
  \vspace{-0.3cm}
  \mbox{\footnotesize(b) $k=9$ @SRN}
\end{minipage}

\begin{minipage}{0.42\linewidth}\centering
  \includegraphics[width=1\textwidth]{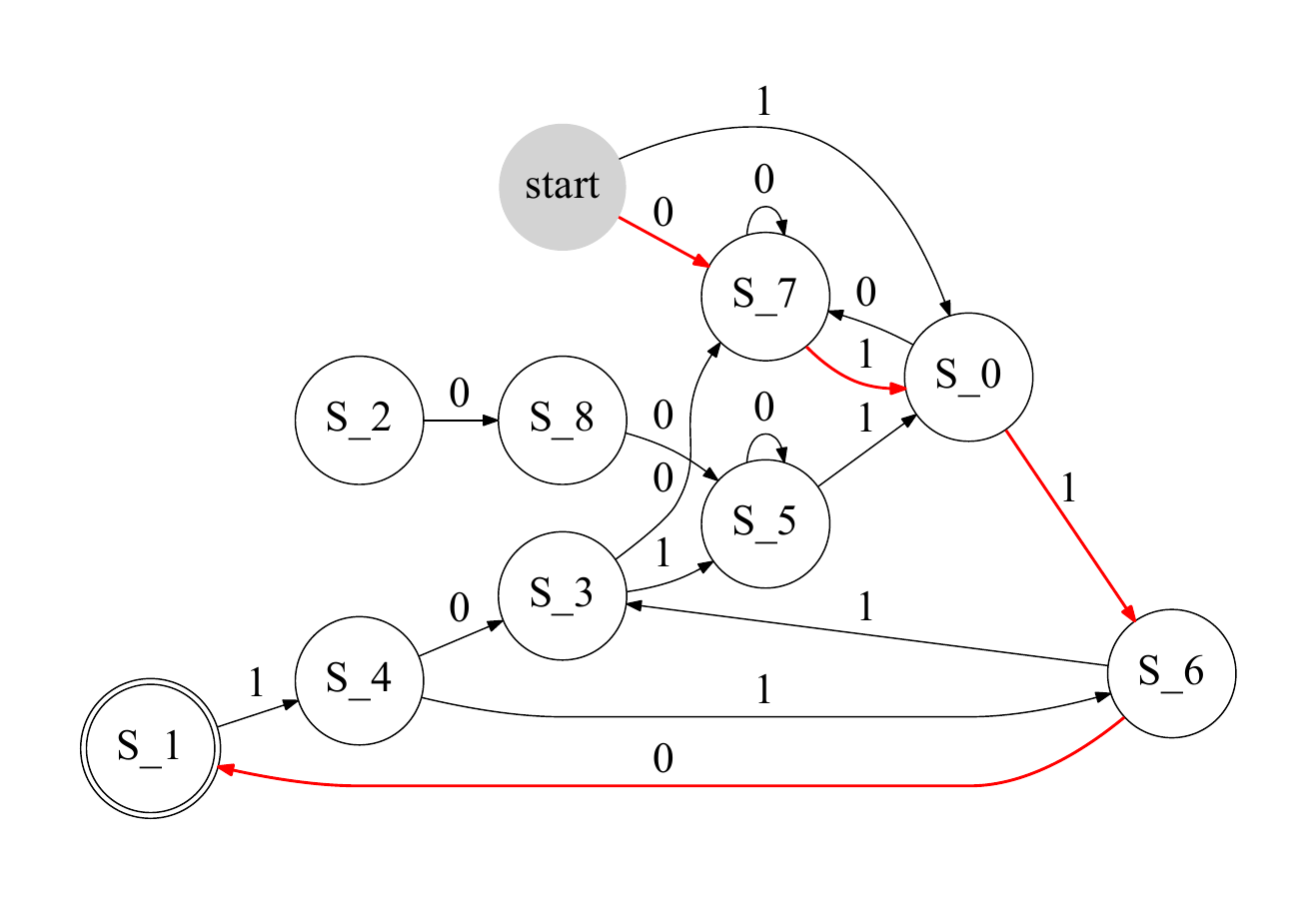}\\
  \vspace{-0.7cm}
  \mbox{\footnotesize(c) $k=9$ @GRU}
\end{minipage}
\begin{minipage}{0.42\linewidth}\centering
  \vspace{0.6cm}  
  \includegraphics[width=1\textwidth]{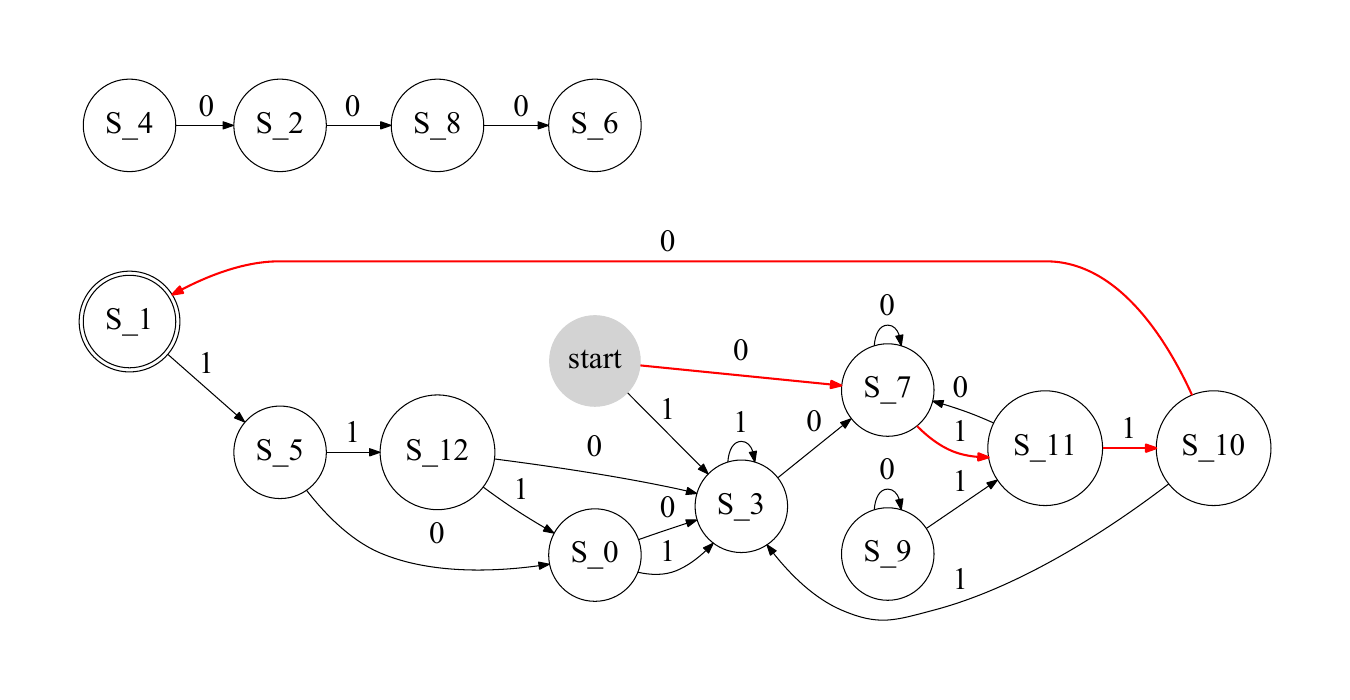}\\
  \vspace{0.1cm}  
  \mbox{\footnotesize(d) $k=13$ @LSTM}
\end{minipage}
\vspace{1em}
\caption{\small FSAs learned from four RNNs trained on task ``0110". The number of clusters $k$ is selected when FSA first reaches accuracy 1.0 as $k$ increases. The 0110 route is marked by red color. Note that in (d) there are four isolated nodes from the main part. This is because we abandon the less frequent transitions when inputting a symbol to learn a deterministic FSA.}
\label{FSA 0110}
\end{figure*}

\begin{figure*}[!t]
\centering 
\small
\begin{minipage}{0.24\linewidth}\centering
  \includegraphics[width=1\textwidth]{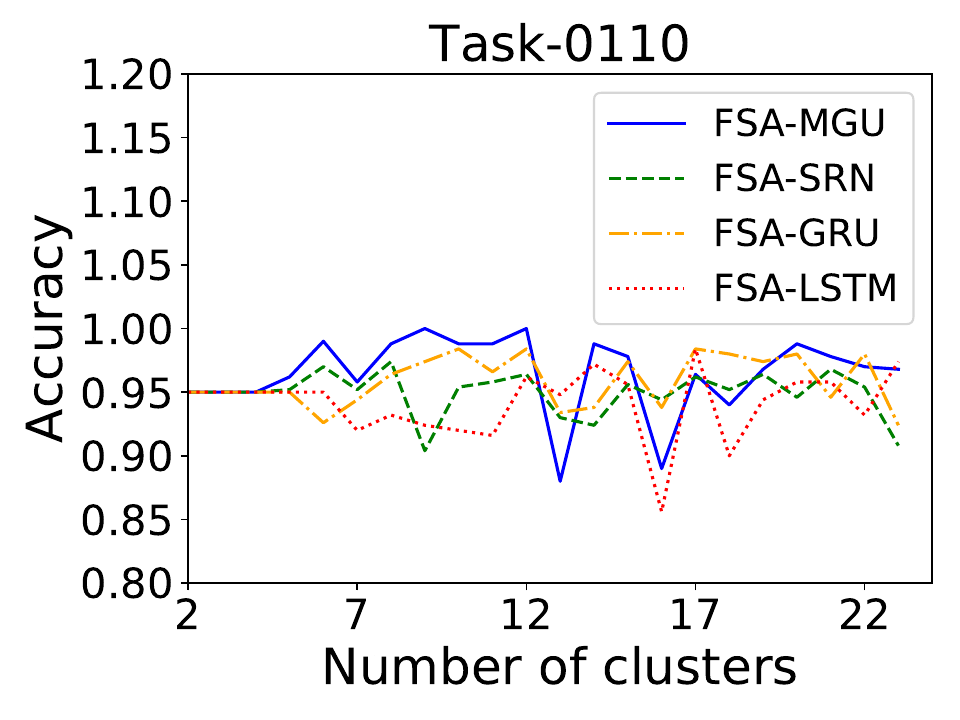}
  \mbox{\footnotesize(a) Identifying 0110 by LISOR-k}
\end{minipage}
\mbox{}\mbox{}
\begin{minipage}{0.24\linewidth}\centering
  \includegraphics[width=1\textwidth]{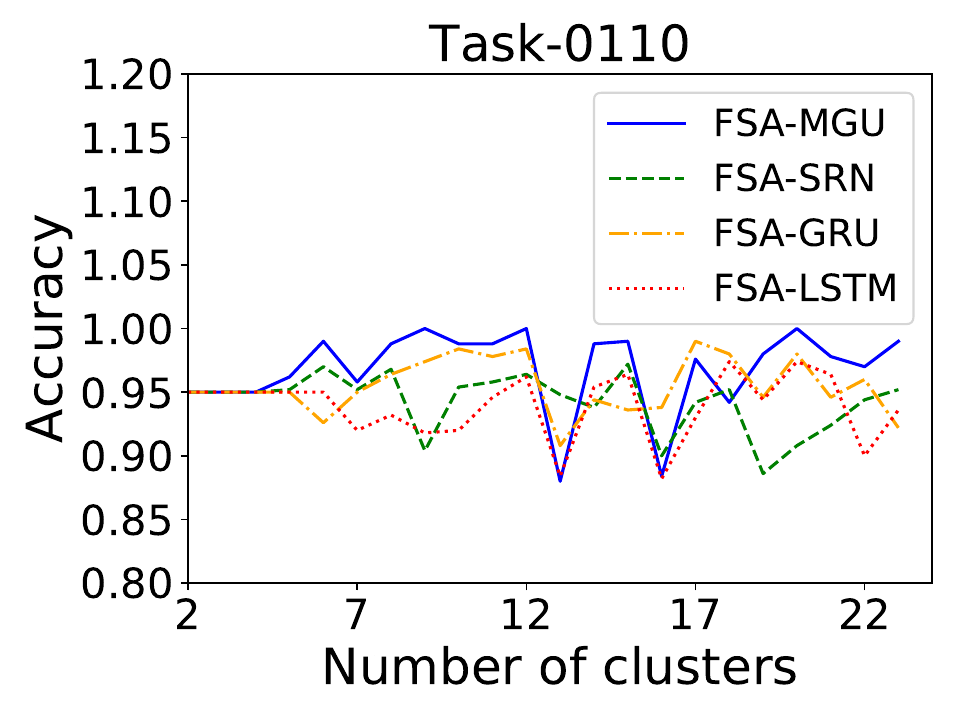}
  \mbox{\footnotesize(b) Identifying 0110 by LISOR-x}
\end{minipage}
\begin{minipage}{0.24\linewidth}\centering
  \includegraphics[width=1\textwidth]{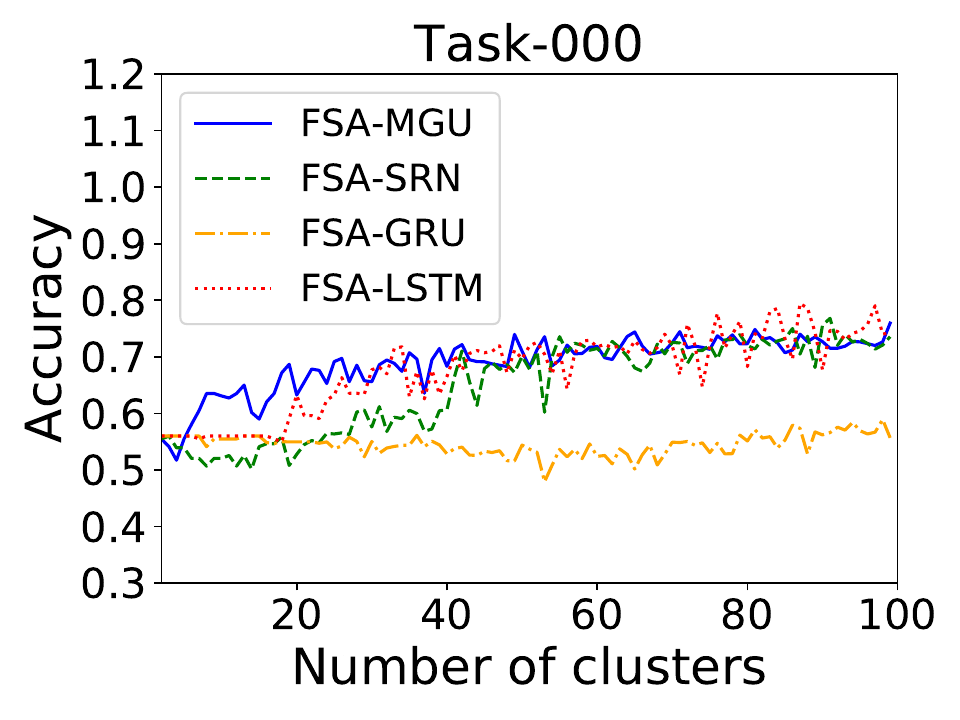}
  \mbox{\footnotesize(c) Identifying 000 by LISOR-k}
\end{minipage}
\mbox{}\mbox{}
\begin{minipage}{0.24\linewidth}\centering
  \includegraphics[width=1\textwidth]{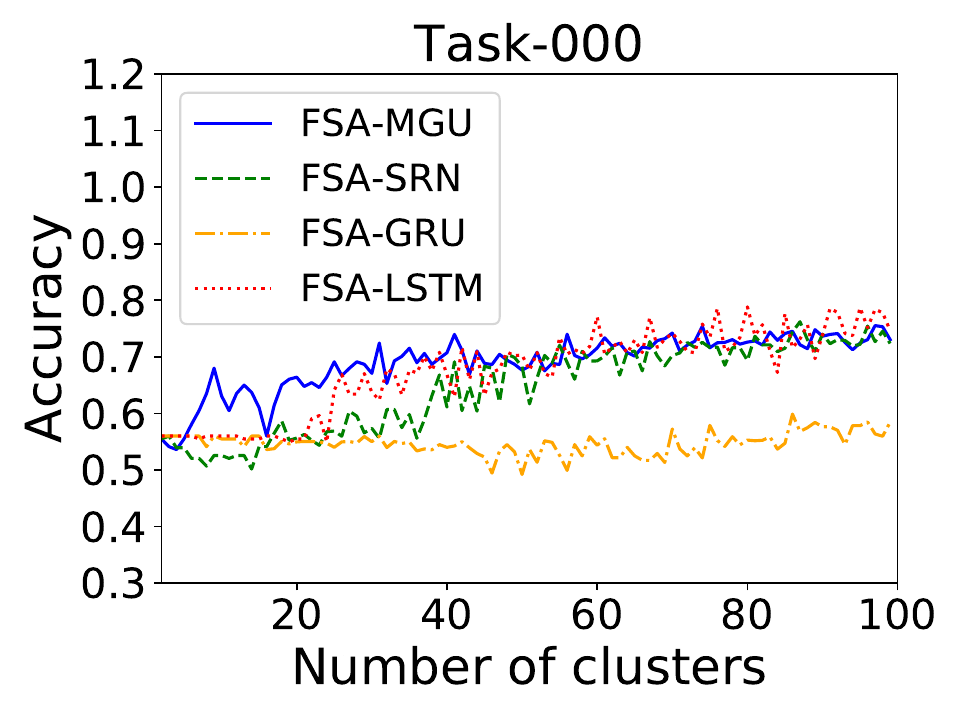}
  \mbox{\footnotesize(d) Identifying 000 by LISOR-x}
\end{minipage}
\caption{\small Impact of the number of clusters on FSA's accuracy when learning from MGU, SRN, GRU and LSTM. We can see that FSA learned from MGU can reach a satisfiable accuracy more quickly.}
\label{figure:num_clusters}
\end{figure*}

In the task ``0110", we set the number of clusters $k$ varying from 2 to 64 (we accumulate $4\times16=64$ hidden points since we only have 16 sequences in validation data and each sequence contains 4 numbers). Table~\ref{table:accuracy of 0110} gives the number of clusters required when the accuracy of FSAs learned from the four RNNs first achieves 1.0 which means perfectly identifying all 0110 sequences. We can see that among all four RNN models, FSA learned from MGU always achieves the accuracy 1.0 with the smallest number of clusters in each trial. Specifically, on average, for LISOR-k the FSA learned from MGU firstly achieves accuracy 1.0 when the number of clusters is 6 followed by that of SRN at cluster number 11. The third one is the FSA learned from GRU with 11.2 clusters, and the final one is that of LSTM with 14.6. For LISOR-x, the corresponding numbers of clusters are 6, 10.6, 19.6 and 25.2, respectively. We can see that the cluster method k-means-x does not bring too many merits on this simple task compared to k-means++. It reduces the number of clusters of FSA learned from SRN but increases those of FSAs learned from GRU and LSTM. This phenomenon can be explained that k-means++ already performs well enough due to the simplicity of this task, and thus k-means-x does not have space to improve.

In the task ``000", we set the number of clusters $k$ ranging from 2 to 200. Actually we have $500\times n$ hidden state points where $n$ is the average length of all the $500$ sequences, but we do not need that many since similar to the task ``0110", large number of clusters may not bring much to performance improvement but may make interpretation from FSA more difficult. This is a more complicated task than task ``0110" and neither the original RNN models nor the learned FSA can reach accuracy 1.0 just like that of task ``0110". Thus, we focus on the accuracy over 0.7, i.e., we will increase the number of clusters until the accuracy of the learned FSA model reaches an accuracy of 0.7. As can be seen from Table~\ref{table:accuracy of 000}, on average for LISOR-k, FSA learned from MGU firstly achieves accuracy over 0.7 when there are 13.6 clusters. Then FSA learned from LSTM achieved this goal with 31.8 clusters followed by that of SRN at cluster number 72.2. The final one is FSA learned from GRU that achieves 0.7 when the number of clusters is 159.4. For LISOR-x, the corresponding numbers of clusters for FSA learned from MGU, SRN, GRU and LSTM are 12.2, 56.2, 157.2 and 34.4, respectively. We can see that cluster method k-means-x plays a role in this task, i.e., lowers the number of clusters of MGU, SRN and GRU.

\begin{figure*}[!t]
	\centering 
	\small
	\begin{minipage}{0.24\linewidth}\centering
	  \includegraphics[width=1\textwidth]{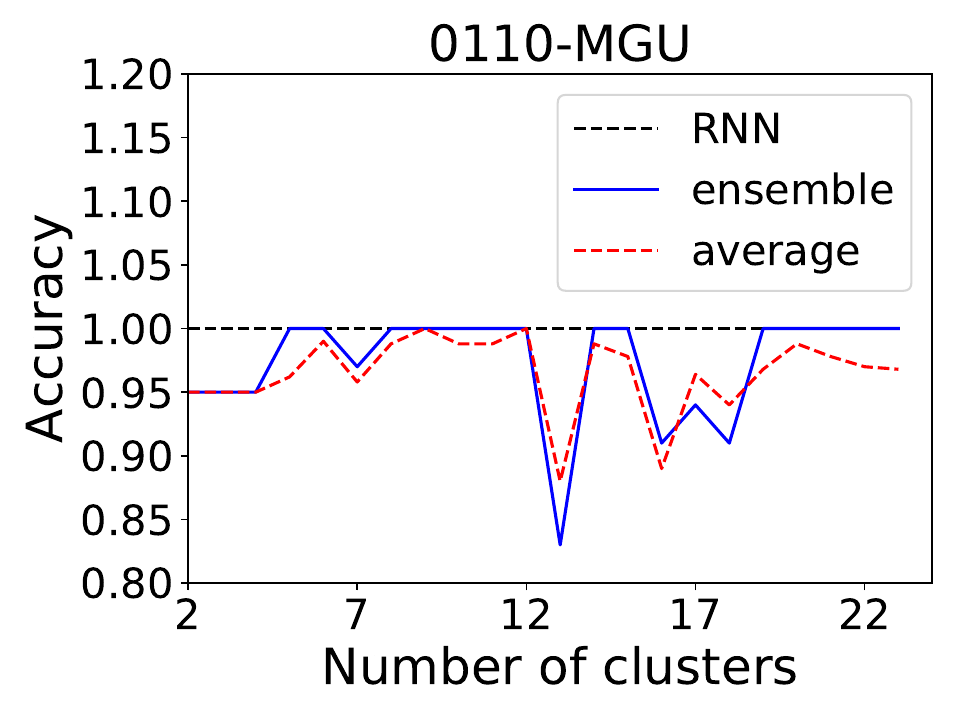}
	  \mbox{\footnotesize(a) Identifying 0110 by LISOR-k}
	\end{minipage}
	\mbox{}\mbox{}
	\begin{minipage}{0.24\linewidth}\centering
	  \includegraphics[width=1\textwidth]{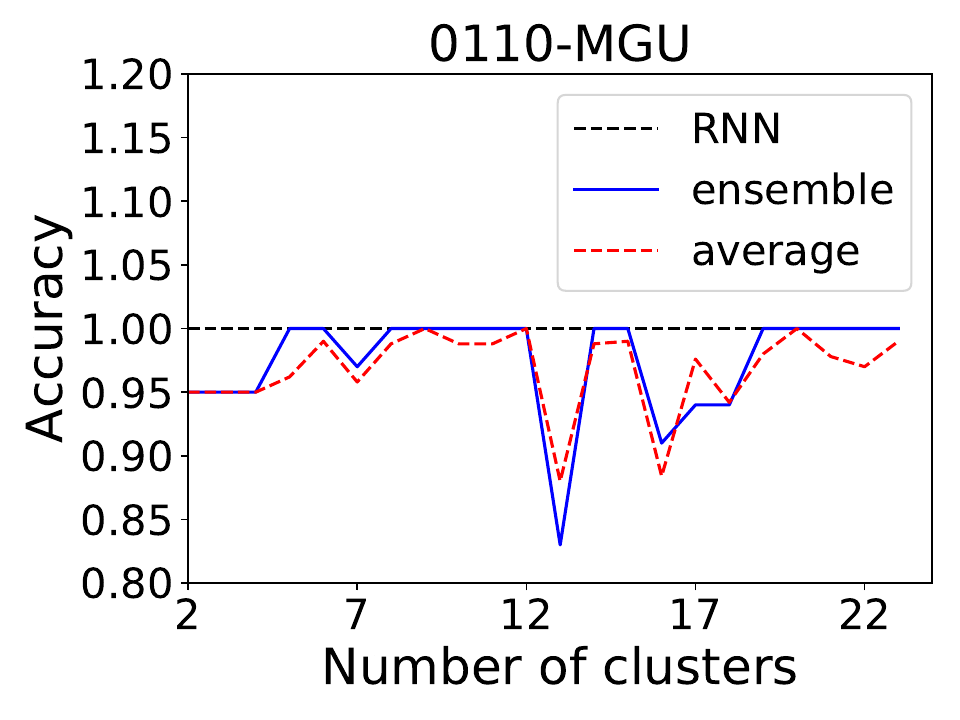}
	  \mbox{\footnotesize(b) Identifying 0110 by LISOR-x}
	\end{minipage}
	\begin{minipage}{0.24\linewidth}\centering
	  \includegraphics[width=1\textwidth]{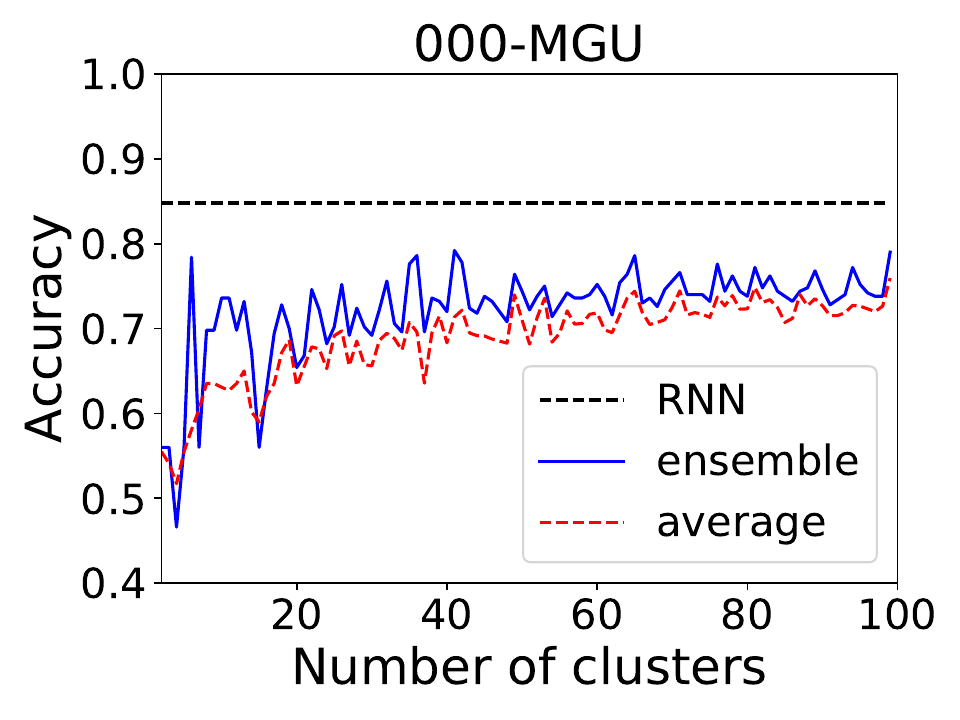}
	  \mbox{\footnotesize(c) Identifying 000 by LISOR-k}
	\end{minipage}
	\mbox{}\mbox{}
	\begin{minipage}{0.24\linewidth}\centering
	  \includegraphics[width=1\textwidth]{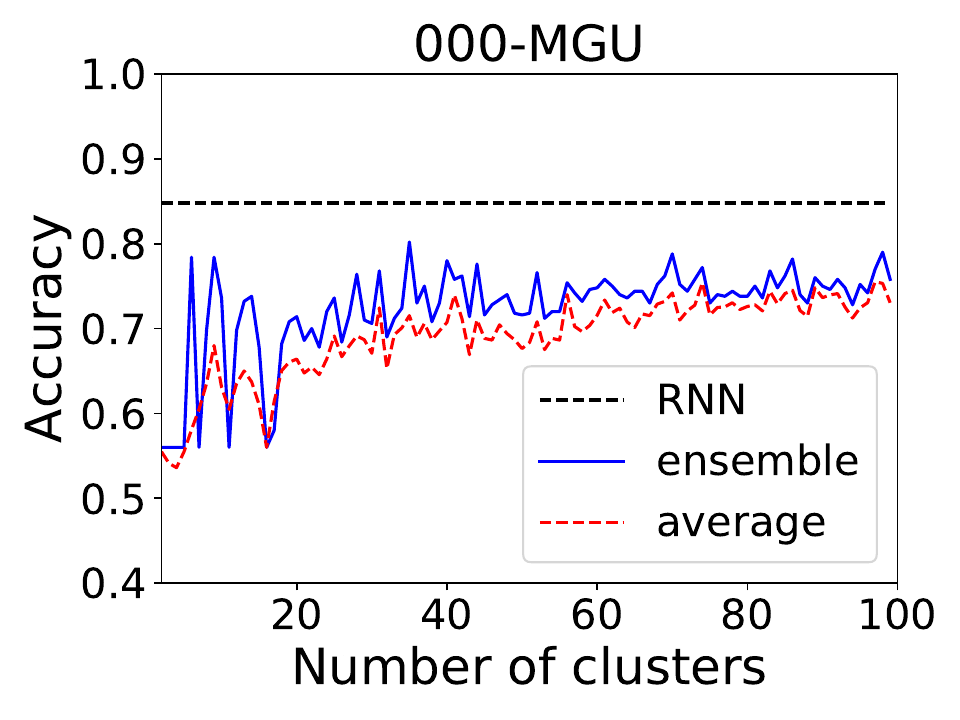}
	  \mbox{\footnotesize(d) Identifying 000 by LISOR-x}
	\end{minipage}
	\caption{\small Comparisons between the average accuracy of the 5 trials and the accuracy of the ensemble of MGU with the increasing of the number of clusters. The black dot line represents the accuracy of the corresponding RNN.}
	\label{figure:ensemble}
	\end{figure*}
	
	\begin{figure*}[!t]
	\centering 
	\small
	\begin{minipage}{0.24\linewidth}\centering
	  \includegraphics[width=1\textwidth]{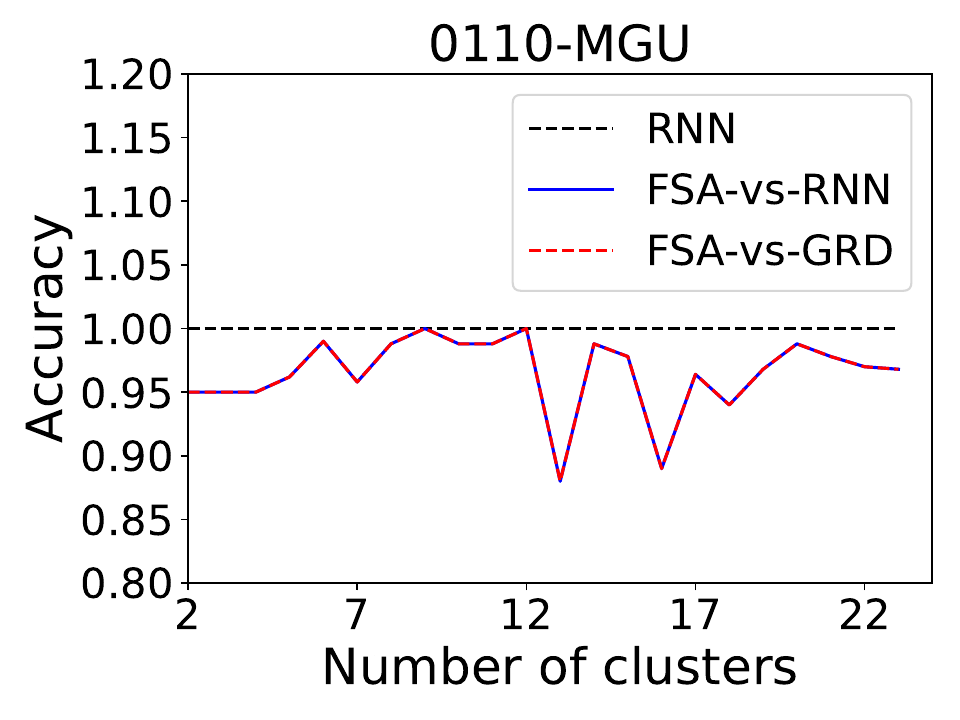}
	  \mbox{\footnotesize(a) Identifying 0110 by LISOR-k}
	\end{minipage}
	\mbox{}\mbox{}
	\begin{minipage}{0.24\linewidth}\centering
	  \includegraphics[width=1\textwidth]{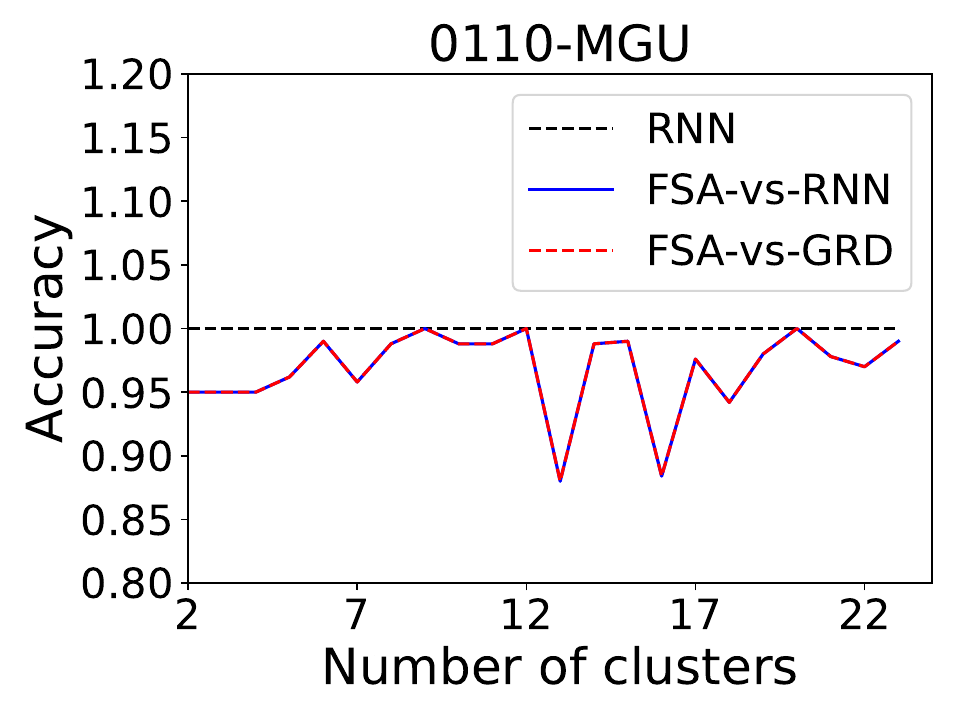}
	  \mbox{\footnotesize(b) Identifying 0110 by LISOR-x}
	\end{minipage}
	\begin{minipage}{0.24\linewidth}\centering
	  \includegraphics[width=1\textwidth]{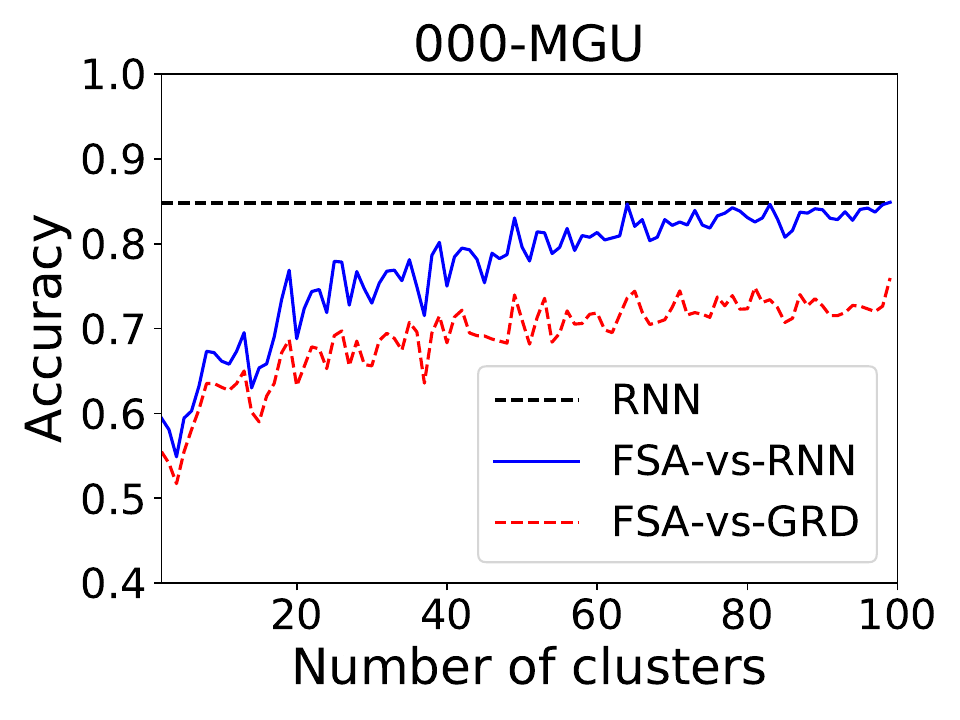}
	  \mbox{\footnotesize(c) Identifying 000 by LISOR-k}
	\end{minipage}
	\mbox{}\mbox{}
	\begin{minipage}{0.24\linewidth}\centering
	  \includegraphics[width=1\textwidth]{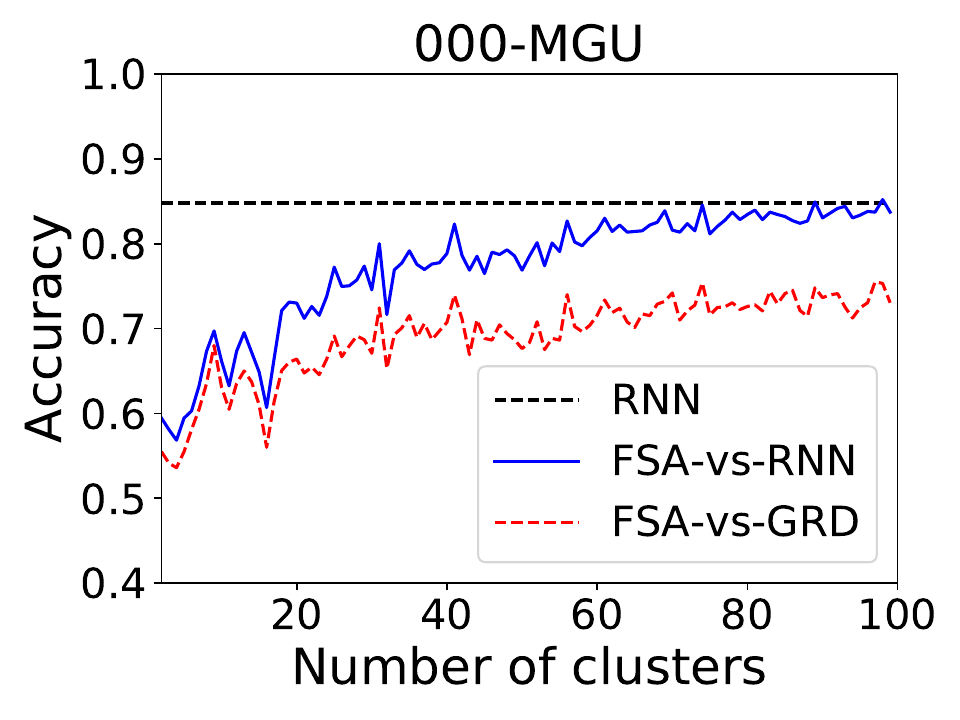}
	  \mbox{\footnotesize(d) Identifying 000 by LISOR-x}
	\end{minipage}
	\caption{\small Comparisons between the average accuracy of the 5 trials versus GRD (abbr. groundtruth) and the average accuracy of the 5 trials versus the output of MGU with the increasing of the number of clusters. The blue and red lines in (a-b) overlap totally because the output of MGU is the same as GRD in task ``0110" due to its simplicity. The black dot line represents the accuracy of the corresponding RNN.}
	\label{figure:vRNN}
	\end{figure*}

\subsubsection{Graphical Illustration of FSA}

In order to visualize the corresponding FSA for each RNN model, we focus on our first method LISOR-k and task ``0110" as an example. LISOR-x and task ``000" have similar results. We choose the number of clusters $k$ that most approaches the average number. For LISOR-k, the average number of $k$ that first achieves accuracy 1.0 for MGU, SRN, GRU and LSTM are 6, 11, 11.2 and 14.6 according to Table~\ref{table:accuracy of 0110}. Thus, we set the number of clusters for MGU to be 6 from trail 3, SRN to be 9 from trial 2, GRU to be 9 from trail 5, LSTM to be 13 from trial 1, respectively. 

We then illustrate FSAs' structure to give people a visual impression of the proposed LISOR's output in Figure~\ref{FSA 0110}, drawn by Graphviz~\cite{DBLP:books/sp/04/EllsonGKNW04}. Here we use gray circle and double circle to represent start and accepting states, respectively. We mark paths of 0110 sequence by red color. As can be seen, for all length-4 zero-one sequences, only 0110 will eventually lead us to an accepting state by following the transitions in illustrated FSAs, and other sequences cannot reach the accepting state. We want to emphasize that by following the flow of FSAs, transitions between states are caused directly by input word. We need not do any numerical calculation that is done in RNN models, thus making the whole process easier to be simulated by human beings. In this way, the learned FSAs are interpretable since they conform to the definition of interpretability mentioned in the first paragraph of Introduction.


\subsubsection{Interpretation about Gate Effect}
We have a first impression in section~\ref{subsubsection: number of clusters} that MGU can achieve guaranteed accuracy with smaller number of clusters. In order to find out the effect of gates, we give a more detailed result, i.e., how the accuracy of the learned FSA changes when the number of clusters is increasing.

For task ``0110", the average accuracy tendencies of five trials are shown in Figure~\ref{figure:num_clusters} (a) and \ref{figure:num_clusters} (b), which correspond to algorithm LISOR-k and LISOR-x, respectively. Here we limit the number of clusters to be less than 24, since when it is larger than 24, the performance changes slightly. In Figure~\ref{figure:num_clusters} (a) and \ref{figure:num_clusters} (b), all FSA models can reach high performance with small number of clusters since the task is not complex. When the number of clusters increases, FSA's performance may be unstable due to the loss of information when we abandon less frequent transitions. We can see that the FSA learned from MGU always firstly achieves high accuracy and holds the lead.

For task ``000", the average accuracy tendencies of five trials until the number of clusters is 100 are shown in Figure~\ref{figure:num_clusters} (c) and \ref{figure:num_clusters} (d). As can be seen from Figure~\ref{figure:num_clusters} (c) and \ref{figure:num_clusters} (d), all four FSAs' accuracies increase with number of clusters increasing. MGU firstly achieves high accuracy and holds the lead. 

In summary, we observe that the FSA learned from MGU reaches its best performance earlier than other RNN models when the number of clusters increases. Therefore, MGU is the most efficient when its learned FSA possesses more clear illustration and easier interpretability. Inspired by this phenomenon together with the fact that MGU contains less gates on the unit than GRU and LSTM, and also the fact that SRN contains no gates, we tend to treat the gate as a regularizer controlling the complexity of the learned FSAs, as well as the complexity of space of hidden state points, while no gate at all will lead to under-fitting. This conclusion motivates us to design other RNN models in the future, which necessarily contain gates, but contain only minimal number of gates.

\subsubsection{Ensemble Results of Multiple FSAs}

Generally, ensemble of multiple classifiers will improve the classification performance~\cite{zhou2012ensemble}. In this section, we will show the ensemble accuracy results with the increasing of number of clusters of the learned FSA. We focus on MGU since its learned FSA outperforms others from the previous experiment results. We train five MGUs with different initializations of parameters. After we got the corresponding FSAs, we give the final output by majority voting, e.g., only when 3 out of 5 FSAs vote for positive, the output will be positive. The results are shown in Figure~\ref{figure:ensemble}. We can see that in both tasks, ensemble of multiple FSAs does improve classification performance. It shows that the ensemble of learned multiple structures will lead to better classification in our tasks. We further observe that on the more complex ``000" task, the improvement is much larger than that on the easier task ``0110". We conjecture that ensemble of multiple FSAs is more suitable for complex tasks and will continue to use this strategy in more complex real tasks. We also provide the classification results of the original RNNs exhibited by black dot lines. As can be seen from Figure~\ref{figure:ensemble}, the performance of FSA is usually worse than the original RNN. This is because for better illustrating, we discard some less frequent transitions to obtain a deterministic FSA in which FSA only transits to a deterministic state when receiving an input. This operation will lose some information and thus makes FSA perform worse than the original RNNs. Nevertheless, what we care about is the consistency between RNN and its learned FSA. Figure~\ref{figure:vRNN} and Table~\ref{table:accuracy of real vRNN} have validated this point. In this way, FSA has a chance to substitute its RNN in applications that need trust. This will be discussed in the next subsection. We also provide the performance of the original RNNs in Figure~\ref{figure:vRNN}, Table~\ref{table:accuracy of real} and Table~\ref{table:accuracy of real vRNN} for complete comparison. In order to avoid duplication, we will not elaborate this comparison in the following.

\subsubsection{The Consistency of FSA Compared to RNN}
\label{section:consistency}
In order to show that the proposed model really reflects the decision process of RNN, or in other words FSA has consistent behavior with its RNN from which it learned, we make the learned FSA compare to the output of RNNs. The similar issue has been discussed in Zhou (2004)~\cite{DBLP:journals/jcst/Zhou04} where the consistency between neural networks and its extracted rule is called \emph{fidelity}. Zhou (2004) put forward the \emph{fidelity-accuracy dilemma}, which highlights that if the extracted rules can even be more accurate than the neural network, then enforcing the fidelity will sacrifice the accuracy of the extracted rules. However, in our case, we do not need to worry about this problem because we are facing more complex tasks instead of merely testing on grammar. Thus our learned FSA has no chance to be more accurate than its RNN due to the loss of information caused by the operation of abandoning the less frequent transitions. Therefore, higher fidelity will bring higher accuracy and thus FSA has a chance to substitute its RNN in applications that need trust. The results are presented in Figure~\ref{figure:vRNN}, which shows the comparisons between accuracy versus groundtruth and the accuracy versus the output of MGU with the increasing of the number of clusters. As can be seen from Figure~\ref{figure:vRNN} (c-d), in task ``000", the accuracy versus the output of MGU is always better than that versus groundtruth. The two lines in Figure~\ref{figure:vRNN} (a-b) overlap totally because the output of MGU is the same as GRD in task ``0110" due to its simplicity. These results all show that FSA has consistent behavior with its RNN from which it learned or in other words, FSA reflects the decision process of RNN, which verifies the strong connection between FSA and its RNN from which it learned. With this fact, it is reasonable to use FSA to probe into RNN and explore the interesting interpretability of RNN itself. The similar results of real task are presented in Table~\ref{table:accuracy of real vRNN}. As can be seen, the accuracies versus RNN for all RNNs are better than the accuracies versus groundtruth. To avoid repetition, we will not confirm this point in the real task. 

\begin{figure*}[!h]
	\setlength{\abovecaptionskip}{1mm}
	\setlength{\belowcaptionskip}{-0.cm}
	\centering 
	\small
	\begin{minipage}{0.49\linewidth}\centering
	  \includegraphics[width=1\textwidth]{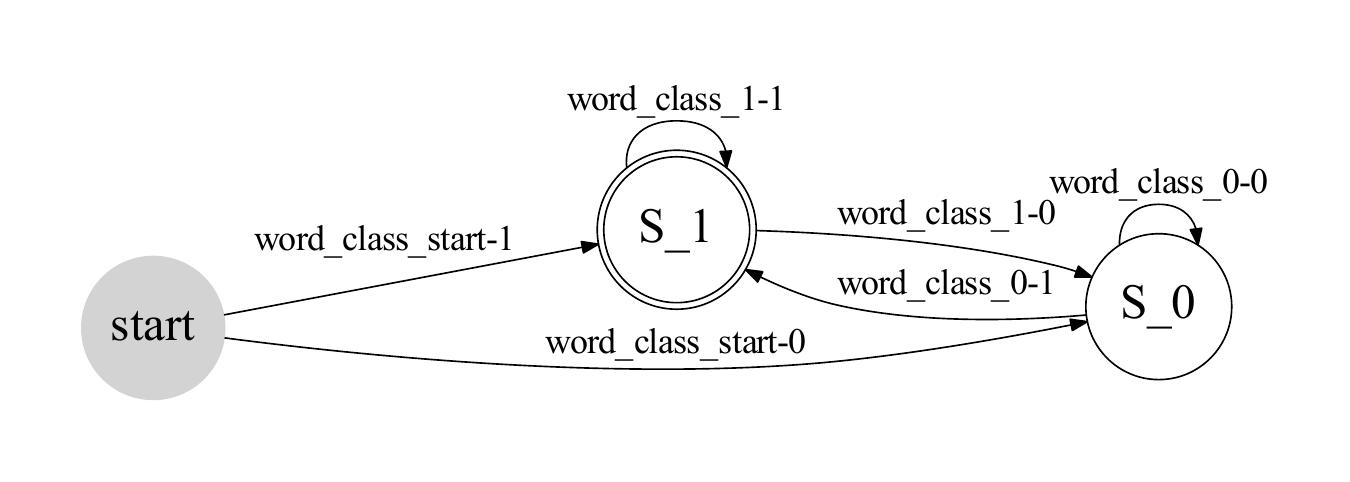}\\
	  \vspace{-0.7cm}
	  \mbox{\footnotesize(a) $k=2$ of MGU from Trail 1 Trail 5}
	\end{minipage}
	\begin{minipage}{0.49\linewidth}\centering
	  \includegraphics[width=1\textwidth]{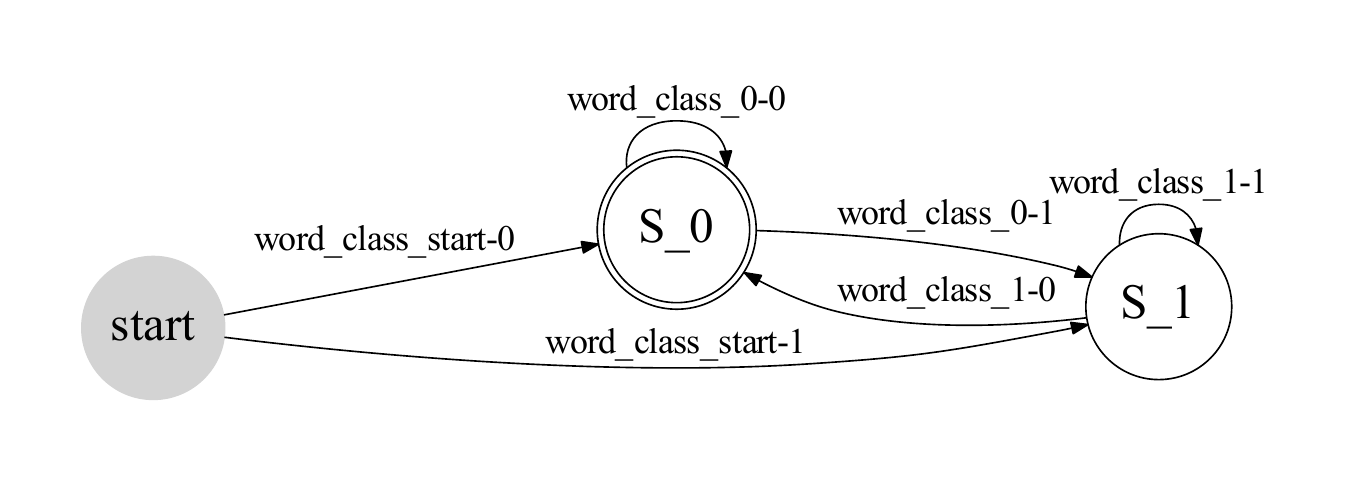}\\
	  \vspace{-0.7cm}
	  \mbox{\footnotesize(b) $k=2$ of MGU from Trail 2 Trail 3 Trail 4}
	\end{minipage}
	\caption{\small FSAs learned from MGU trained on sentiment analysis task. The FSA here is shrunk and the words between the same two states with same direction are grouped into a word class. For example, the words in ``word\_class\_0-1" all incur transitions from State 0 (S\_0) to State 1 (S\_1). We train five MGUs with different initializations. (a) is the result of trial 1 and trial 5 where the accepting state is State 1. (b) is the result of trial 2, 3 and 4 where the accepting state is State 0.}
	\label{FSA sentiment}
	\end{figure*}

\subsection{Real Task}
In this section, we conduct our experiments on a practical task about sentiment analysis.

\subsubsection{Settings}


In this task, we will use the IMDB dataset~\cite{DBLP:conf/acl/MaasDPHNP11} to do sentiment analysis~\cite{DBLP:conf/acl/MaasDPHNP11,DBLP:conf/acl/PoriaCHMZM17}, which is a very common task in natural language processing. In this dataset, each instance is the comment for a movie and the task is to classify the given sentence into positive or negative sentiment. 

To train the RNN models, we first use word2vec~\cite{DBLP:journals/corr/abs-1301-3781} to map each English word from film reviews into a 300 dimensions numerical vector. Then we train four different RNNs (MGU, SRN, GRU and LSTM) using these vectors as input. All RNNs' dimension of hidden states and number of hidden layers are set to be 10 and 3 respectively, and we randomly select 2000 random-length film reviews as training data. After we get the trained RNNs, we learn FSAs using 200 testing data. Note that we adopt a transductive setting, i.e. using the test data directly to learn FSAs to ensure all words in test data's vocabulary are fully covered. In order to explore multiple structures and obtain stable results, for each of MGU, SRN, GRU and LSTM, we train five different ones according to different initializations and learn five corresponding FSAs from them.

\begin{table*}[!t]
	\begin{minipage}[!t]{\columnwidth}
	\renewcommand\arraystretch{1.2}
	\setlength{\abovecaptionskip}{0.cm}
	\setlength{\belowcaptionskip}{-0.cm}
		\caption{\small Accuracy on sentiment analysis task when the number of cluster is 2. ``Average" means the average accuracy results of the five structured FSAs. ``Ensemble" means using ensemble technique to combine the five structured FSAs to improve the performance. LISOR-k and LISOR-x are our methods. In each method and each strategy, the highest accuracy is bold among the four RNNs.}
		\label{table:accuracy of real}
		\vspace{0.8mm}
		\centering
		\small
		\setlength\tabcolsep{1.5pt}
	\setlength\tabcolsep{3pt}
			\begin{tabular}{c|c|c|c|c|c}
			
	\hline
	\multirow{2}{*}{RNN Type} & \multirow{2}{*}{RNN Acc} & \multicolumn{2}{c|}{LISOR-k} & \multicolumn{2}{c}{LISOR-x}\\
		\cline{3-6}  &  & Average & Ensemble & Average & Ensemble \\
		\hline 
		MGU &0.818& \textbf{0.701} & 0.740 & \textbf{0.740} & \textbf{0.850}\\
		SRN &0.647& 0.604 & 0.635 & 0.592 & 0.615\\
		GRU &0.811& 0.662 & 0.660 & 0.699 & 0.780\\
		LSTM &0.720& 0.669 & \textbf{0.750} & 0.669 & 0.755\\
	
	\hline
			\end{tabular}
	\end{minipage}
	\quad
	\begin{minipage}[!t]{\columnwidth}
	\renewcommand\arraystretch{1.2}
	\setlength{\abovecaptionskip}{0.cm}
	\setlength{\belowcaptionskip}{-0.cm}
		\caption{\small Accuracy on sentiment analysis task when the number of cluster is 2. ``vGRT" means the average accuracy results of the five structured FSAs versus groundtruth. ``vRNN" means the average accuracy results of the five structured FSAs versus the output of RNN. LISOR-k and LISOR-x are our methods. In each method and each RNN, the higher accuracy is bold between vGRT and vRNN.}
		\label{table:accuracy of real vRNN}
		\vspace{0.8mm}
		\centering
		\small
		\setlength\tabcolsep{6.5pt}
			\begin{tabular}{c|c|c|c|c|c}
			
	\hline
	\multirow{2}{*}{RNN Type} & \multirow{2}{*}{RNN Acc} & \multicolumn{2}{c|}{LISOR-k} & \multicolumn{2}{c}{LISOR-x}\\
		\cline{3-6} &   & vGRT & vRNN & vGRT & vRNN \\
		\hline 
		MGU &0.818& 0.701 & \textbf{0.749} & 0.740 & \textbf{0.778}\\
		SRN &0.647& 0.604 & \textbf{0.729} & 0.592 & \textbf{0.719}\\
		GRU &0.811& 0.662 & \textbf{0.721} & 0.699 & \textbf{0.746}\\
		LSTM &0.720& 0.669 & \textbf{0.751} & 0.669 & \textbf{0.753}\\
	
	\hline
			\end{tabular}
	\end{minipage}
	\end{table*}
	
	
	
	\begin{table*}[!t]
		\renewcommand\arraystretch{1.5}
		\setlength{\abovecaptionskip}{0.cm}
		\setlength{\belowcaptionskip}{-0.cm}
			\caption{\small The word class (word\_class\_0-1) leading transition from State 0 to State 1 (the accepting state) contains mainly ``positive" words. Here the number in the bracket shows the serial number of the FSA from which this word comes.
			}
			\label{table:group1}
			\vspace{0.8mm}
			\centering
			\small
				\begin{tabular}{l|l} 
		\hline
		 Positive&
		\begin{minipage}{0.85\textwidth}
		\vspace{0.1cm}
		exceptionally(1) riffs(1) Wonderful(1) gratitude(1) diligent(1) spectacular(1) sweetness(1) exceptional(1) Best(1) feats(1) sexy(1) bravery(1) beautifully(1) immediacy(1) meditative(1) captures(1) incredible(1) virtues(1) excellent(1) shone(1) honor(1) pleasantly(1) lovingly(1) exhilarating(1) devotion(1) teaming(1) humanity(1) graceful(1) tribute(1) peaking(1) insightful(1) frenetic(1) romping(1) proudly(1) terrific(1) Haunting(1) sophisticated(1) strives(1) exemplary(1) favorite(1) professionalism(1) enjoyable(1) alluring(1) entertaining(1) Truly(1) noble(1) bravest(1) exciting(1) Hurray(1) wonderful(1) Miracle(1)... feelings(5) honest(5) nifty(5) smashes(5) ordered(5) revisit(5) moneyed(5) flamboyance(5) reliable(5) strongest(5) loving(5) useful(5) fascinated(5) carefree(5) recommend(5) Greatest(5) legendary(5) increasing(5) loyalty(5) respectable(5) clearer(5) priority(5) Hongsheng(5) notable(5) reminiscent(5) spiriting(5) astonishing(5) charismatic(5) lived(5) engaging(5) blues(5) pleased(5) subtly(5) versatile(5) favorites(5) remarkably(5) poignant(5) Breaking(5) heroics(5) promised(5) elite(5) confident(5) underrated(5) justice(5) glowing(5) ... adventure(1,5) victory(1,5) popular(1,5) adoring(1,5) perfect(1,5) mesmerizing(1,5) fascinating(1,5) extraordinary(1,5) AMAZING(1,5) timeless(1,5) delight(1,5) GREAT(1,5) nicely(1,5) awesome(1,5) fantastic(1,5) flawless(1,5) beguiling(1,5) famed(1,5)
		\vspace{0.1cm}
		\end{minipage}\\
		\hline
		Negative&
		\begin{minipage}{0.85\textwidth}
		\vspace{0.1cm}
		downbeat(1) wicked(1) jailed(1) corruption(1) eccentric(5) troubled(5) cheats(5) coaxed(5) convicted(5) steals(5) painful(5) cocky(5) endures(5) annoyingly(5) dissonance(5) disturbing(5) goofiness(1,5)
		\vspace{0.1cm}
		\end{minipage}\\
		\hline
				\end{tabular}
				
		\end{table*}
		
		
		
		\begin{table*}[!t]
		\renewcommand\arraystretch{1.1}
		\setlength{\abovecaptionskip}{0.cm}
		\setlength{\belowcaptionskip}{-0.cm}
			\caption{\small The word class (word\_class\_0-1) leading transition from State 0 to State 1 (the rejecting state) contains mainly ``negative" words. Here the number in the bracket shows the serial number of the FSA from which this word comes.
			}
			\label{table:group2}
			\vspace{0.8mm}
			\centering
			\small
				\begin{tabular}{l|l}
		\hline
		 Positive&
		\begin{minipage}{0.85\textwidth}
		\vspace{0.1cm}
		merry(2) advance(2) beliefs(3) romancing(3) deeper(3) resurrect(3) 
		\vspace{0.1cm}
		\end{minipage}\\
		\hline
		Negative&
		\begin{minipage}{0.85\textwidth}
		\vspace{0.1cm}
		shut(2) dullest(2) unattractive(2) Nothing(2) adulterous(2) stinkers(2) drunken(2) hurt(2) rigid(2) unable(2) confusing(2) risky(2) mediocre(2) nonexistent(2) idles(2) horrible(2) disobeys(2) bother(2) scoff(2) interminably(2) arrogance(2) mislead(2) filthy(2) dependent(2) MISSED(2) asleep(2) unfortunate(2) criticized(2) weary(2) corrupt(2) jeopardized(2) drivel(2) scraps(2) phony(2) prohibited(2) foolish(2) reluctant(2) Ironically(2) fell(2) escape(2) ... whitewash(3) fanciful(3) flawed(3) No(3) corrupts(3) fools(3) limited(3) missing(3) pretense(3) drugs(3) irrational(3) cheesy(3) crappy(3) cheap(3) wandering(3) forced(3) warped(3) shoplift(3) concerns(3) intentional(3) Desperately(3) dying(3) clich(3) bad(3) evil(3) evicted(3) dead(3) minor(3) drunk(3) loser(3) bothered(3) reek(3) tampered(3) inconsistencies(3) ignoring(3) Ward(3) doom(3) quit(3) goofier(3) antithesis(3) fake(3) helplessness(3) surly(3) demoted(3) fault(3) worst(3) baffling(3) destroy(3) fails(3) Pity(3) pressure(3) nuisance(3) farce(3) fail(3) worse(3) SPOLIER(3) egomaniacal(3) quandary(3) burning(3) drinker(3) blame(3) intimidated(3) perfidy(3) boring(3) conservative(3) forgetting(3) hostile(3) ... unattractive(2,3) goof(2,3) lousy(2,3) stupidest(2,3) mediocrity(2,3) Badly(2,3) mediocre(2,3) waste(2,3) hypocrite(2,3) confused(2,3) vague(2,3) clumsily(2,3) stupid(2,3)
		\vspace{0.1cm}
		\end{minipage}
		\\
		\hline
				\end{tabular}
				
		\end{table*}

\subsubsection{Discussion on the Number of Clusters}
Note that this task is more complex than the artificial tasks, thus we cannot enumerate over all possible number of clusters (i.e., number of hidden states in RNNs). We have tried different number of clusters, that is $k$, from $2$ to $20$ and found that the smaller $k$ is, the better the performance. We understand that if the number of clusters is large enough, FSA will perform better and have similar performance with corresponding RNN models. However, when $k$ is small, our empirical results show that simple structure may lead to better performance. Thus in this part, we only exhibit the results when the number of clusters is 2. In this case, all the FSAs possess the simplest structure that is easy to simulate and understand as well as be visually illustrated. With same number of clusters, the FSA with higher accuracy has more advantage. 

\subsubsection{Graphical Illustration of FSA}
This task has much larger vocabulary size containing thousands of English words, which means the number of symbols (i.e., words) in $\Sigma$ is not simply 2 that is adopted in the artificial tasks. Thus in order to show the graphical illustration of FSA, we shrink the edges with same direction between two states into one edge and illustrate the resulted FSA learned from MGU with two clusters in Figure~\ref{FSA sentiment}. Other FSAs' structures are similar and we omit them. In this way the words on a shrunk edge are naturally grouped into a class named as ``word\_class". Besides, we find that the structures of the five trials are the same but with different accepting state. As can be seen from Figure~\ref{FSA sentiment}, the accepting state of trial 1 and trial 5 are State 1~(S\_1) while those of trail 2, 3 and 4 are State 0~(S\_0).

\subsubsection{Accuracy Result}
We show the average results of the five FSAs' accuracy in Table~\ref{table:accuracy of real} for each RNN. We can see that, for both LISOR-k and LISOR-x, FSAs learned from MGU have the highest accuracy compared to other three RNNs and LISOR-x performs better than LISOR-k, which shows the effectiveness of k-means-x that utilizes the extra position feature. In order to explore the multiple structures, we adopt the same strategy as artificial tasks, i.e., combining the results of the five FSAs by ensemble using majority voting. The ensemble classification results of FSAs learned from MGU, SRN, GRU and LSTM are also shown in Table~\ref{table:accuracy of real}. As can be seen, for LISOR-k, the results of ensemble method are almost better than the case without ensemble except GRU and FSA learned from MGU exhibits competitive performance. For LISOR-x, the performances of ensemble are all better than the cases without ensemble and the FSA learned from MGU outperforms other RNNs' FSAs. LISOR-x is better than LISOR-k in MGU, GRU and LSTM as well. Table~\ref{table:accuracy of real vRNN} shows the consistency between RNN and its learned FSA. To avoid repetition, for more details, please refer to Section~\ref{section:consistency}.

\subsubsection{Semantic Interpretation}
We find that the FSA learned from RNN gives semantic aggregated states and its transition graph shows us a very interesting vision of how RNNs intrinsically handle text classification tasks.
Specifically, we still focus on MGU since its FSA possesses best performance. The results are shown in Table~\ref{table:group1} and Table~\ref{table:group2}. We consider the transition from State 0 to State 1 in all the five learned FSAs. Table~\ref{table:group1} shows the results on the 1st and 5th FSA. According to Figure~\ref{FSA sentiment} (a), we realize that this is a transition leading to the accepting state. We can see that the word class leading transition from State 0 to State 1 (the accepting state) contains mainly ``positive" words, for example, wonderful, spectacular, sweetness, etc. The results on the 2nd and 3rd FSA are shown in Table~\ref{table:group2}. According to Figure~\ref{FSA sentiment} (b), we know State 1 is a rejecting state. We can see that most of the activation words of this transition are negative, for example, dullest, unattractive, confusing, etc. We also make exploration on the effect of multiple structures. The number in the bracket in Table~\ref{table:group1} and Table~\ref{table:group2} shows the serial number of the FSA from which this word comes. As can be seen, one FSA will only cover one part of the positive or negative words, thus having limited semantic meaning, while multiple FSAs can make the semantic meaning more plentiful.

\section{Conclusion}

It will be beneficial if we can learn an interpretable structure from the RNN models since there is still no clear understanding of the inner mechanism of RNN models. In this paper, realizing the similarity between RNNs and FSA, as well as the good interpretability of FSA, we try to learn FSA from RNN, and analyze RNNs from FSA's point of view. After verifying that the hidden states of gated RNNs indeed form clusters, we propose two methods to learn FSAs from four kinds of RNNs, based on different clustering strategies. We graphically show the learned FSA and explicitly give the transition route for human beings to simulate. We also show how the number of gates affects the performance of RNNs, and the semantic meaning behind the numerical calculation in hidden units. We find that gate is important for RNN but the less the better, which can guide us to design other RNNs with only one gate and save us from unnecessary twists and turns. Considering FSA can consistently imitate the performance of its RNN from which it learned, we would like to devote to improving the performance of RNN and FSA. In this way, FSA can have a chance to substitute its RNN from which it learned to be used in those applications that need trust, such as the applications involving human being’s lives or dangerous facilities.

Despite the contributions summarized in the last paragraph, we want to convey an insight to our community. That is it is feasible and desirable to find other interpretable tools that resemble the complex deep learning models to interpret them. For example, we find that FSA is analogous to RNNs, and then we employ FSA to probe into RNNs to discover their interpretable mechanisms. However, this work is still a preliminary exploration and it has many aspects to improve. For instance, due to the limitation of FSA's ability on binary classification, the most appropriate real task that we can perform is the sentiment analysis task since it is a binary classification task. We would like to find or propose other tools that is with interpretability but can handle complex tasks rather than binary classification in the future. Another interesting future issue is to incorporate some approach like the proposed one into the recently proposed \textit{abductive learning}~\cite{Zhou2019}, a new paradigm which encompasses machine learning and logical reasoning, such that the machine learning results become more comprehensible to enable a more powerful technology when combining with logical reasoning. 



\section*{Acknowledgement} 
This research was supported by the National Key R\&D Program of China (2018YFB1004300), the National Science Foundation of China (61751306, 61921006), and the Collaborative Innovation Center of Novel Software Technology and Industrialization.

\bibliographystyle{IEEEtran}{
\bibliography{Reference}}

\begin{IEEEbiography}[{\includegraphics[width=1in,height=1.25in,clip,keepaspectratio]{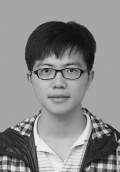}}]{Bo-Jian Hou}
is a PhD student in the Department of Computer Science \& Technology of Nanjing University. He received the BSc degree from Nanjing University, China, in 2014. He is currently working toward the PhD degree in computer science at Nanjing University. His main research interests include machine learning and data mining. He won the National Scholarship in 2017. He also won the Program A for Outstanding PhD Candidate of Nanjing University and CCFAI Outstanding Student Paper Award in 2019.
\end{IEEEbiography}

\begin{IEEEbiography}[{\includegraphics[height=1.25in,clip,keepaspectratio]{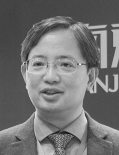}}]{Zhi-Hua Zhou} (S'00-M'01-SM'06-F'13) received the BSc, MSc and PhD degrees in computer science from Nanjing University, China, in 1996, 1998 and 2000, respectively, all with the highest honors. He joined the Department of Computer Science \& Technology at Nanjing University as an Assistant Professor in 2001, and is currently Professor, Head of the Department of Computer Science and Technology, and Dean of the School of Artificial Intelligence; he is also the Founding Director of the LAMDA group. His research interests are mainly in artificial intelligence, machine learning and data mining. He has authored the books \textit{Ensemble Methods: Foundations and Algorithms}, \textit{Evolutionary Learning: Advances in Theories and Algorithms}, \textit{Machine Learning} (in Chinese), and published more than 200 papers in top-tier international journals or conference proceedings. He has received various awards/honors including the National Natural Science Award of China, the IEEE Computer Society Edward J. McCluskey Technical Achievement Award, the ACML Distinguished Contribution Award, the PAKDD Distinguished Contribution Award, the IEEE ICDM Outstanding Service Award, the Microsoft Professorship Award, etc. He also holds 24 patents. He is the Editor-in-Chief of the \textit{Frontiers of Computer Science}, Associate Editor-in-Chief of the \textit{Science China Information Sciences}, Action or Associate Editor of the \textit{Machine Learning}, \textit{IEEE Transactions on Pattern Analysis and Machine Intelligence}, \textit{ACM Transactions on Knowledge Discovery from Data}, etc. He served as Associate Editor-in-Chief for \textit{Chinese Science Bulletin} (2008-2014), Associate Editor for \textit{IEEE Transactions on Knowledge and Data Engineering} (2008-2012), \textit{IEEE Transactions on Neural Networks and Learning Systems} (2014-2017), \textit{ACM Transactions on Intelligent Systems and Technology} (2009-2017), \textit{Neural Networks} (2014-2016), etc. He founded ACML (Asian Conference on Machine Learning), served as Advisory Committee member for IJCAI (2015-2016), Steering Committee member for ICDM, ACML, PAKDD and PRICAI, and Chair of various conferences such as Program Chair of AAAI 2019, General Chair of ICDM 2016, and Area Chair of NeurIPS, ICML, AAAI, IJCAI, KDD, etc. He was the Chair of the IEEE CIS Data Mining Technical Committee (2015-2016), the Chair of the CCF-AI (2012-2019), and the Chair of the CAAI Machine Learning Technical Committee (2006-2015). He is a foreign member of the Academy of Europe, and a Fellow of the ACM, AAAI, AAAS, IEEE, IAPR, IET/IEE, CCF, and CAAI.
\end{IEEEbiography}

\end{document}